\documentclass{article}

\usepackage[preprint]{neurips_2026}

\usepackage[utf8]{inputenc} 
\usepackage[T1]{fontenc}    
\usepackage{hyperref}       
\usepackage{url}            
\usepackage{booktabs}       
\usepackage{amsfonts}       
\usepackage{nicefrac}       
\usepackage{microtype}      
\usepackage{xcolor}         
\usepackage{multirow}
\usepackage{caption}
\usepackage{subcaption}
\usepackage{graphicx}
\usepackage{tikz}
\usetikzlibrary{arrows.meta, positioning, calc}
\usepackage{amssymb}
\usepackage[dvipsnames, table]{xcolor}
\usepackage{amsmath}
\usepackage{wrapfig}
\usepackage{enumitem}
\usepackage{tabularx}
\usepackage{nicematrix}
\usepackage{booktabs}
\usepackage[table]{xcolor}
\usepackage{nicematrix}
\usepackage{booktabs}
\usepackage{xfp}
\definecolor{MyGreen}{HTML}{F1F8E9} 
\definecolor{DeepGreen}{HTML}{A5D6A7} 
\definecolor{MyRed}{HTML}{FFEBEE}   
\definecolor{DeepRed}{HTML}{EF9A9A}

\newcommand{\AutoColor}[3]{%
    \edef\diff{\fpeval{abs(#1 - #2)}}%
    \edef\intensity{\fpeval{min(80, \diff * 20)}}
    \ifstr{#3}{pos}{
        \ifdim \fpeval{#1}pt > \fpeval{#2}pt
            \cellcolor{DeepGreen!\intensity!MyGreen} #1
        \else
            \ifdim \fpeval{#1}pt < \fpeval{#2}pt
                \cellcolor{DeepRed!\intensity!MyRed} #1
            \else
                #1 
            \fi
        \fi
    }{
        
        \edef\val{\fpeval{strip_unit(substitute_k(#1))}} 
        \edef\refval{\fpeval{strip_unit(substitute_k(#2))}}
        \ifdim \val pt < \refval pt
            \cellcolor{DeepGreen!\intensity!MyGreen} #1
        \else
            \cellcolor{DeepRed!\intensity!MyRed} #1
        \fi
    }%
}
\usepackage{arydshln}

\usepackage{algorithm}
\usepackage{algpseudocode}
\usepackage[most]{tcolorbox}
\usepackage{xcolor}

\title{Reasoning Compression with Mixed-Policy Distillation}

\author{
Han Yang\textsuperscript{1,2}\thanks{Equal contribution.} \quad Mingyan Wu\textsuperscript{3$*$}
\quad\textbf{Bailan He}\textsuperscript{4,5}
\quad \textbf{Zeyu Cao}\textsuperscript{6}
\quad \textbf{Sikuan Yan}\textsuperscript{4}\\
\quad \textbf{Kevin Qinghong Lin\textsuperscript{7}}
\quad \textbf{Zifeng Ding\textsuperscript{6,8$*$}}\thanks{Corresponding author.}\\
\textsuperscript{1}Technical University of Munich, 
\textsuperscript{2}GESIS – Leibniz Institute for the Social Sciences,\\
\textsuperscript{3}Northeastern University, 
\textsuperscript{4}LMU Munich,
\textsuperscript{5}Siemens AG,\\
\textsuperscript{6}University of Cambridge,
\textsuperscript{7}University of Oxford,
\textsuperscript{8}Mina AI\\
\texttt{zifeng@mina-ai.co}
\\
}

\begin{document}

\maketitle

\begin{abstract}
Reasoning-centric large language models (LLMs) achieve strong performance by generating intermediate reasoning trajectories, but often incur excessive token usage and high inference-time decoding cost. We observe that, when solving the same problems, larger reasoning models can often produce more concise traces, whereas smaller reasoning models tend to generate longer and more redundant trajectories. This is especially problematic in real-world deployment, where memory, latency, and serving-cost constraints often favor smaller models. Our observations suggest that reasoning compression can be transferred from large models to small ones rather than enforced through explicit length constraints. Based on this insight, we propose Mixed-Policy Distillation (MPD), a reasoning compression framework that transfers concise reasoning behavior from a larger-sized teacher to a smaller student by distilling teacher-compressed student trajectories. Unlike on-policy distillation, which aligns the student with teacher distributions over verbose student trajectories, or off-policy distillation, which relies on teacher-generated trajectories and may suffer from distribution mismatch, MPD combines the strengths of both. Given a student-sampled trajectory, the teacher rewrites it into a more concise reasoning trace, and the student is trained via KL-based alignment on the compressed trajectory. This preserves student-policy exploration while injecting teacher-guided compression. Experiments on Qwen3-1.7B show that MPD reduces token usage by up to 27.1\% while improving performance across multiple reasoning benchmarks, demonstrating an effective approach to efficient small-model reasoning.
\end{abstract}

\section{Introduction}

LLMs exhibit improved reasoning performance when equipped with explicit chain-of-thought (CoT) traces~\citep{wei2022cot, kojima2022zeroshot}. Recent reasoning-centric models, such as OpenAI~o1, DeepSeek-R1\citep{deepseekr1}, and newer systems like GPT-5.5 and DeepSeek-V4, push this paradigm further by generating intermediate reasoning trajectories that often span tens of thousands of tokens per query. While such extended reasoning improves accuracy, it also incurs substantial inference-time decoding cost, since computation scales with the length of the generated trace. This motivates the need for new paradigms that improve reasoning efficiency by reducing token usage, i.e., through reasoning compression.

\begin{figure}[t]
    \centering
    \begin{subfigure}{0.48\linewidth}
        \centering
        \includegraphics[width=\linewidth]{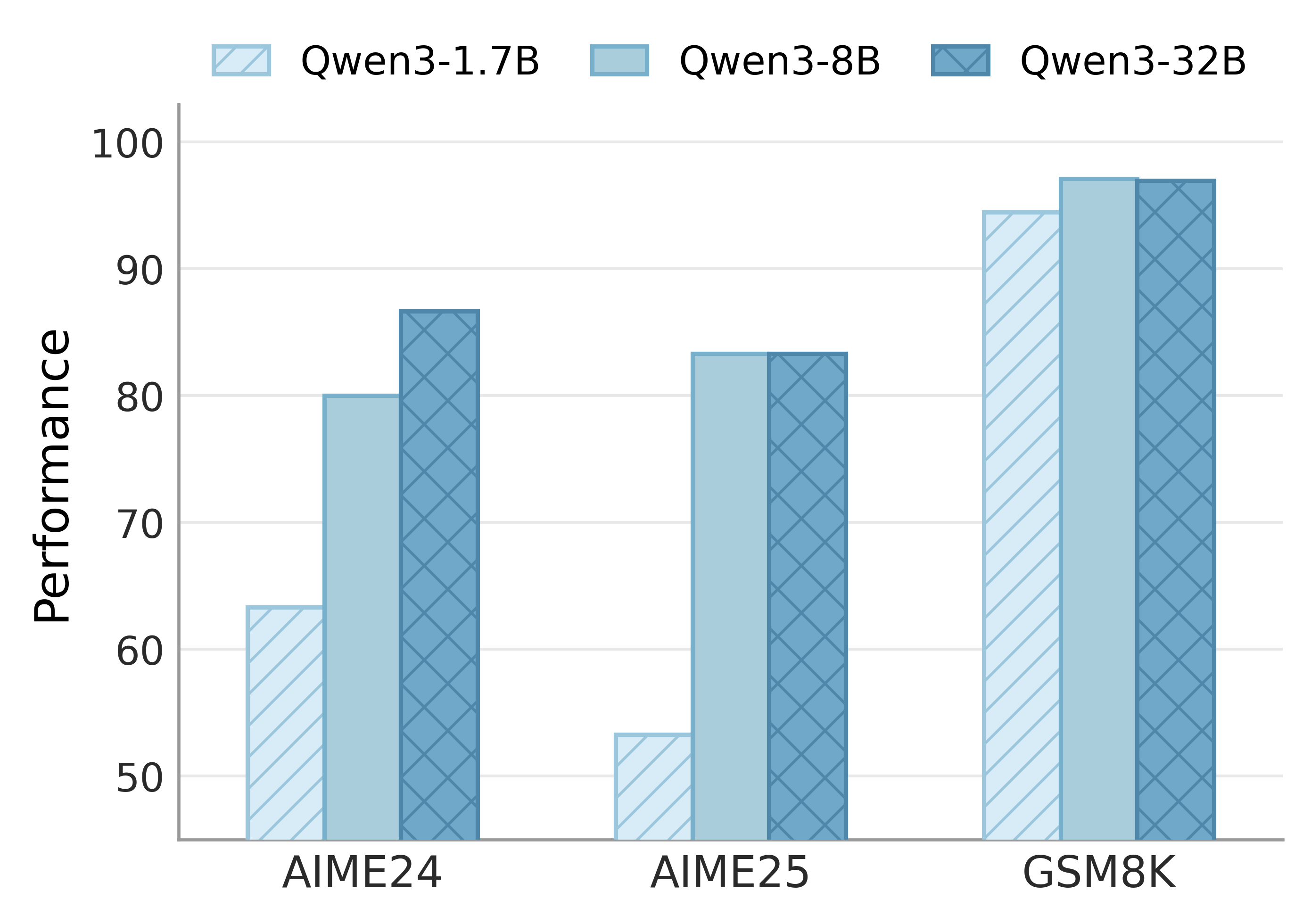}
        \caption{Performance.}
        \label{fig:performance}
    \end{subfigure}
    \hfill
    \begin{subfigure}{0.48\linewidth}
        \centering
        \includegraphics[width=\linewidth]{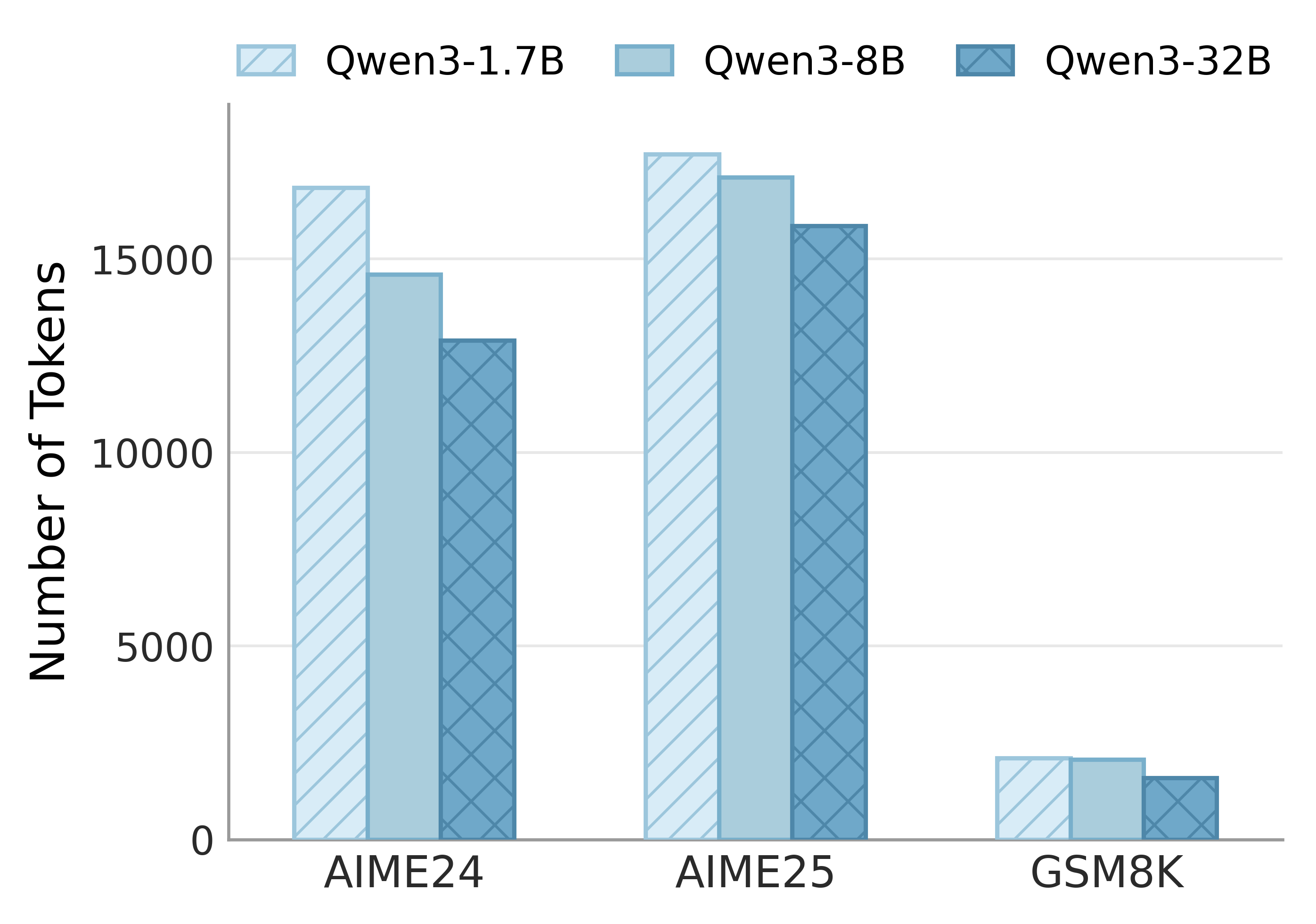}
        \caption{Token Consumption.}
        \label{fig:token}
    \end{subfigure}
    \caption{Reasoning performance (a) and token usage (b) across models of different sizes from the same family. Evaluated on GSM8K, AIME24, and AIME25 using Pass@4 as the evaluation metric.}\label{fig:preliminary_performance_token}
    \vspace{-15pt}
\end{figure}

Reasoning inefficiency is especially pronounced in smaller models. We observe that, when solving the same reasoning problems, smaller reasoning models tend to produce longer traces with more redundant intermediate steps, whereas larger models often reach correct answers with substantially fewer tokens. For example, Qwen3-32B reduces token usage by around 23\% compared with Qwen3-1.7B on the same problems (Fig. \ref{fig:preliminary_performance_token}). This creates a practical tension: real-world deployment often favors smaller models because of memory, latency, and serving-cost constraints, yet these are precisely the models that suffer most from redundant reasoning. At the same time, the concision of larger models suggests a natural source of supervision: concise reasoning behavior can be transferred from a larger model to a smaller one, rather than imposed through explicit length penalties.


Existing reasoning compression methods mainly fall into three categories. Training-free methods modify prompts or decoding strategies to elicit shorter traces~\citep{sui2025stop, wang2025wait,DBLP:conf/fllm/RenzeG24,DBLP:journals/corr/abs-2502-18600}, but are prompt-sensitive and provide only shallow control over reasoning length. SFT- and RL-based methods directly optimise models for concise reasoning: SFT trains on externally shortened traces~\citep{DBLP:conf/aaai/KangSCZ25, xia2025tokenskip,litecot}, creating a train-inference discrepancy because the student does not learn from its own reasoning behaviour, while RL~\citep{jin2025recut, DBLP:journals/corr/abs-2402-03300,aggarwal2025l1,DBLP:journals/tmlr/HouZJLQAC26} penalises length through reward design, often requiring ground-truth answers and risking suppression of useful exploration. More recently, CRISP~\citep{opsdc_crisp} applies on-policy self-distillation to train on the student’s own generations and avoid distribution shift. However, it relies on a prompted variant of the student itself rather than actual generations from a stronger teacher, limiting compression quality.

In this paper, we propose \textbf{Mixed-Policy Distillation} (MPD), a framework that combines on-policy student generation with external-teacher compression. Given a reasoning trajectory generated by the student, a larger-sized teacher rewrites it into a more concise version, and the student is trained to align with the rewritten trajectory via token-level KL divergence. This creates a mixed-policy signal: the trajectory structure comes from the student, while the compressed content comes from the teacher. Unlike previous methods, MPD preserves the student’s own problem-solving strategy while using the teacher to compress and refine that strategy into a more concise reasoning trace. MPD is designed for real-world deployment settings, where smaller models are often preferred, and reasoning compression is most needed for these less capable yet more practical models.

We summarize our contributions as follows:
\begin{itemize}[leftmargin=1.4em, labelsep=0.4em, itemsep=0pt, topsep=1pt]
    \item We propose Mixed-Policy Distillation, a reasoning compression framework that aligns a student with teacher-compressed versions of its own reasoning trajectories, preserving the student’s problem-solving strategy while using a larger-sized teacher to enable concise reasoning.    
    \item We instantiate MPD on OpenMathInstruct-2~\citep{openmath} with Qwen3-1.7B~\citep{yang2025qwen3} as the student and Qwen3-8B as the teacher, training without ground-truth supervision, reward models, or length penalties. On in-domain mathematical reasoning, MPD reduces inference tokens by
    27.1\% on Qwen3-1.7B while maintaining or improving accuracy.    
    \item We further evaluate MPD on out-of-domain reasoning tasks, including code reasoning and long-context reasoning. MPD consistently reduces token usage with improved performance, demonstrating that the compressed reasoning behavior generalizes beyond the training domain.
\end{itemize}

\section{Related Work and Preliminaries}

\subsection{Related Work}
We discuss notable related work on reasoning trace compression for reasoning-centric LLMs.
\paragraph{Training-Free Reasoning Compression.}
Large Language Models (LLMs) have demonstrated strong reasoning ability on complex tasks such as mathematics, code generation, and long-context reasoning~\citep{deepseekr1, jaech2024openai, cobbe2021training, nguyen2025codemmlu, bai2025longbench}. However, reasoning-centric LLMs often suffer from overthinking, producing unnecessarily verbose reasoning trajectories that increase decoding cost and latency~\citep{chen2024not, sui2025stop, gekhman2026thinking}. A direct way to reduce reasoning length is to prompt models to produce only essential intermediate steps~\citep{sui2025stop, wang2025wait}. Concise Chain-of-Thought~\citep{DBLP:conf/fllm/RenzeG24} encourages shorter CoT traces, while Chain-of-Draft~\citep{DBLP:journals/corr/abs-2502-18600} asks models to capture only key information. Although simple and broadly applicable, these methods are sensitive to prompt design and remain far from the theoretical efficiency frontier~\citep{lee2025well}. 

\paragraph{SFT and RL-Based Compression.}
Another line of work directly optimises models for concise reasoning. SFT-based methods train LLMs on shortened reasoning trajectories~\citep{DBLP:conf/aaai/KangSCZ25}. TokenSkip~\citep{xia2025tokenskip} learns controllable CoT compression by omitting non-essential tokens, while LiteCoT~\citep{litecot} fine-tunes models on pre-refined concise traces. However, because these compression targets are constructed externally, SFT introduces a train-inference discrepancy: the student learns from curated traces rather than its own reasoning behaviour, and is therefore not explicitly trained to compress or recover from the trajectories it will produce at inference time. RL-based methods instead encourage concise reasoning through length penalties or token budgets~\citep{jin2025recut, DBLP:journals/corr/abs-2402-03300}. L1~\citep{aggarwal2025l1} jointly optimises accuracy and token efficiency, while THINKPRUNE~\citep{DBLP:journals/tmlr/HouZJLQAC26} trains long-thinking models under a strict token budget. Although effective in reducing reasoning length, these methods often require ground-truth answers to define correctness rewards and may suppress useful exploratory reasoning by directly penalising length. To address these limitations, MPD learns concise reasoning without ground-truth answers, externally curated compression targets, or manually designed length rewards, using the student’s own reasoning traces as the basis for training.

\paragraph{Distillation for Concise Reasoning.}
Distillation offers a natural way to learn reasoning compression by transferring efficient reasoning behaviour from teacher models to students. Off-policy distillation trains the student on teacher-generated reasoning trajectories, which can provide concise and high-quality traces, but may suffer from distribution mismatch because these trajectories do not come from the student’s own policy. On-policy distillation instead trains on student-generated trajectories, reducing exposure bias and better matching inference-time reasoning behaviour~\citep{agarwal2024gkd, gu2024minillm, thinkingmachines2024distillation, DBLP:journals/corr/abs-2602-12275}. A recent work, i.e., CRISP~\citep{opsdc_crisp}, applies on-policy self-distillation to reasoning compression by inducing conciseness with explicit brevity instructions. However, it does not leverage the natural compression advantage of larger models. Motivated by our observation that larger models often solve the same problems with substantially shorter traces (Fig.~\ref{fig:preliminary_performance_token}), MPD further turns the teacher into a trajectory compressor, allowing concise reasoning behaviour to be transferred while keeping training grounded in the student’s own inference-time distribution.

\subsection{Preliminaries: Off-Policy vs. On-Policy Distillation}
We formalize off-policy and on-policy distillation for reasoning trajectory learning. We consider transferring reasoning behaviour from a stronger teacher model $\pi_t(\cdot \mid x; \phi)$ to a weaker student model $\pi_s(\cdot \mid x; \theta)$ with learnable parameters $\theta$, where $\phi$ denotes the fixed teacher parameters. Given a question $x \sim D$, let $y=(y_1,\ldots,y_n)$ denote a reasoning trajectory together with the final answer. Distillation trains the student to match the teacher's predictive behaviour along such trajectories, typically by minimizing a divergence $\operatorname{Div}(\cdot\|\cdot)$ between the student and teacher distributions.

A key distinction between off-policy and on-policy distillation is the source of the training trajectories. Off-policy distillation trains on trajectories sampled from the teacher, yielding the objective
\begin{equation}
\mathcal L_{\mathrm{off}} =
\mathbb E_{x \sim D,\; y \sim \pi_t(\cdot \mid x; \phi)}
\left[
\sum_{i=1}^{n}
\operatorname{Div}\left(
\pi_s(\cdot \mid x, y_{<i}; \theta)
\;\|\;
\pi_t(\cdot \mid x, y_{<i}; \phi)
\right)
\right].
\end{equation}
Teacher-generated trajectories are often of higher quality, but they create a train-inference mismatch: the student is optimized under teacher-induced contexts rather than the contexts produced by its own generations~\citep{thinkingmachines2024distillation}. Consequently, the student is not trained to recover from its own mistakes. At inference time, the student must reason over unseen questions under its own generated contexts. Since off-policy distillation trains it only on teacher-generated trajectories, the student is not explicitly exposed to the kinds of imperfect intermediate states it may produce itself, making recovery difficult and allowing errors to accumulate across reasoning steps.


In contrast, on-policy distillation samples trajectories from the student model:
\begin{equation}
\label{formal_opd}
\mathcal L_{\mathrm{on}} =
\mathbb E_{x \sim D,\; y \sim \pi_s(\cdot \mid x; \theta)}
\left[
\sum_{i=1}^{n}
\operatorname{Div}\left(
\pi_s(\cdot \mid x, y_{<i}; \theta)
\;\|\;
\pi_t(\cdot \mid x, y_{<i}; \phi)
\right)
\right].
\end{equation}
This brings a key advantage: the student is aligned with the teacher under its own trajectory distribution. Instead of learning only from teacher-induced contexts, the student is trained on the reasoning states it is likely to encounter during inference. This reduces the mismatch between training and inference, making the distillation objective better aligned with the student’s actual inference-time behaviour.

In the context of reasoning compression, a teacher should be a model that reasons more efficiently, while the student is more prone to redundant and inefficient intermediate reasoning. Rooting from our observation in Fig.~\ref{fig:preliminary_performance_token}, we argue that a smaller reasoning model can be naturally supervised by its larger variant within the same model family, since the larger model provides useful signals for both task performance and token efficiency. We next introduce MPD, a distillation-based framework that mixes on- and off-policy learning: it samples trajectories from the student to preserve the student’s inference-time distribution, while using the larger teacher to compress these trajectories into concise reasoning traces. We then show in Sec.~\ref{sec:results} with empirical evidence to explain why this mixed-policy design outperforms both standard on-policy and off-policy distillation.

\section{Reasoning Compression with Mixed-Policy Distillation}
\label{sec:method}
\begin{figure}[htbp]
    \centering
    \includegraphics[width=\linewidth]{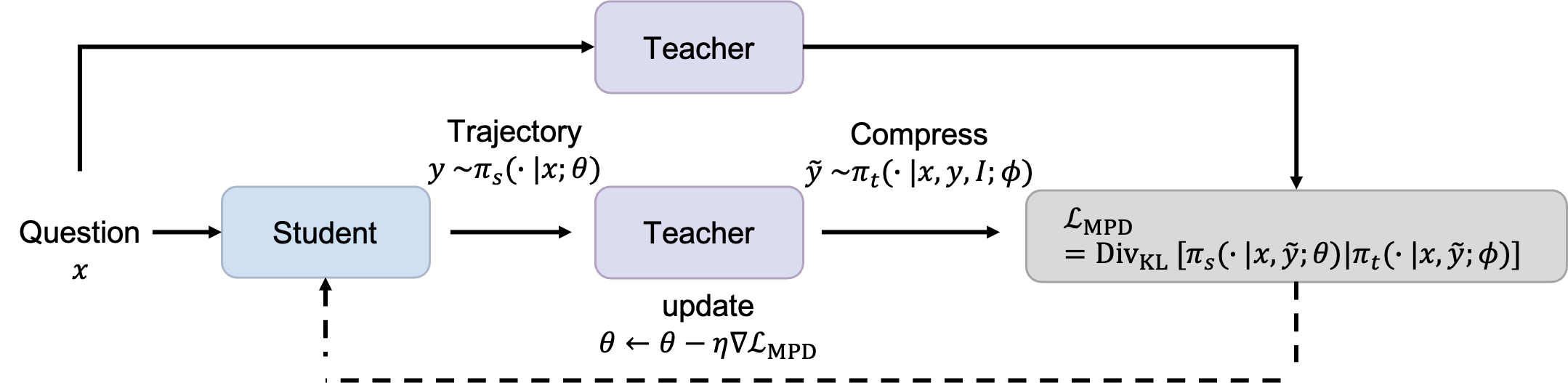}
    \caption{Overview of MPD.}
    \label{fig:method_major}
\end{figure}

We provide the overview of our proposed MPD in Fig.~\ref{fig:method_major}. MPD consists of three stages. First, given a question $x \sim D$, the student model samples a reasoning trajectory with the final answer, $y \sim \pi_s(\cdot \mid x; \theta)$. Second, the larger teacher model compresses this student-generated trajectory into a compressed, semantically equivalent trace, denoted as $\tilde{y}=(\tilde{y}_1,\ldots,\tilde{y}_m)$. Third, the student is trained on the compressed trajectory by minimizing the divergence between the student and teacher distributions conditioned on each prefix $\tilde{y}_{<i}$, using $\pi_t(\cdot \mid x,\tilde{y}_{<i};\phi)$ as the distillation target.

Motivated by our observation that larger models often use substantially fewer tokens than smaller students when solving the same problems, MPD uses a larger teacher as the source of reasoning compression. Specifically, as shown in Table~\ref{table:prompt_teacher}, the teacher is given the same question $x$ together with the student-generated reasoning trace $y$, and is instructed to rewrite the trace into a more concise form.
The compression prompt $I$ is designed to preserve the student’s reasoning signal while allowing the teacher to improve its expression. We ask the teacher to remain faithful to the original student-generated reasoning, avoid introducing new solution paths or additional information, and keep only the steps necessary for answering the question. In this way, the compressed trajectory is not an independently generated teacher solution. Instead, \textbf{it is a teacher-compressed version of the student’s own trajectory, which preserves the student’s problem-solving path and therefore keeps the training signal tied to the student policy.}
At the same time, the compression process is performed by the larger teacher under its own policy. \textbf{When deciding which intermediate steps are necessary, which redundant parts can be removed, and how the remaining reasoning should be rewritten, the teacher injects its stronger ability to organise and express reasoning concisely.} Therefore, the resulting compressed trace $\tilde{y}$ combines two desirable properties: it remains aligned with the student-generated trajectory $y$, while reflecting the teacher’s stronger compression capability. This is the core mechanism that allows MPD to retain the distributional benefit of student-generated trajectories while transferring concise reasoning behaviour from a larger model.

\begin{table}[t]
\centering
\caption{Compression prompt $I$ for the teacher model.} 
\label{table:prompt_teacher}
\small
\begin{tabularx}{\columnwidth}{|X|}
\hline
\rowcolor[gray]{0.9} \textbf{Compression Prompt for Teacher} \\ \hline
Based on the provided reasoning, write a summary. \\
\\
The summary should integrate all relevant information from the original text that can help answer the specified question and form a coherent paragraph. Ensure that the summary includes all useful information from the original text for answering the question. Do not add new information. \\
\\
Question: \{$x$\} \\
Reasoning: \{$y$\} \\
\\
Please provide the summary: \\ \hline
\end{tabularx}
\vspace{-10pt}
\end{table}

To verify whether teacher-compressed reasoning traces preserve the student’s reasoning strategy, we randomly sample 100 questions from OpenMathInstruct-2~\citep{openmath}. For each question, we use Qwen3-1.7B as the student to generate an initial reasoning trajectory, and Qwen3-8B as the teacher to compress it. We then compute sentence embeddings for both the original student trajectory and the teacher-compressed trajectory using Sentence-BERT~\citep{sentencebert}, and measure their cosine similarity. As a reference, we also compute the cosine similarity between each student-generated trajectory and another student-generated trajectory from a different randomly selected question within the same set of 100 questions.
To further isolate the effect of using a larger teacher, we repeat the same compression procedure using also Qwen3-1.7B with the same compression prompt, and report the corresponding embedding similarity. In both teacher-compression and student-compression settings, we also record the compression rate (relative reduction in token usage). This allows us to evaluate whether compression preserves the semantic content of the student’s reasoning while reducing trajectory length. 

From Table~\ref{tab:compression_stats}, we can observe that our compression prompt substantially shortens the student-
\begin{wraptable}{r}{6.5cm}
\vspace{-10pt}
\centering
\small
\caption{Compression effect using different sizes of models with our prompt. }
\label{tab:compression_stats}
\resizebox{\linewidth}{!}{%
\begin{tabular}{lrrrr}
\toprule
\textbf{Compressor} & \textbf{Comp. Rate} & \textbf{Sim. w. Init} & \textbf{Sim. w. Rnd} \\
\midrule
Qwen3-1.7B & 87.5\% & 0.8102 & 0.1660 \\
Qwen3-8B & 93.0\% & 0.8098 & 0.1659 \\
\bottomrule
\end{tabular}
}
\vspace{-10pt}
\end{wraptable}
generated reasoning traces while preserving their semantic content. Qwen3-8B achieves a higher compression rate (Comp. Rate) than Qwen3-1.7B, reducing the trace length by 93.0\% compared with 87.5\%, which verifies that larger models are better at expressing reasoning concisely, even when the student is identical to the compressor in the Qwen3-1.7B compression setting. Meanwhile, both compressed traces remain highly similar to the initial student trajectory, with cosine similarity (Sim. w. Init) around 0.81, while their similarity to random trajectories (Sim. w. Rnd) is only around 0.16. This verifies that our compression prompt does not push the compressed trace far away from the original reasoning signal, but mainly removes redundant steps.

After obtaining the compressed trajectory $\tilde{y}$, we perform distillation on this trajectory. The MPD training objective is therefore defined as:
\begin{equation}
\mathcal L_{\mathrm{MPD}} =
\mathbb E_{x \sim D,\;
y \sim \pi_s(\cdot \mid x; \theta),\;
\tilde{y} \sim \pi_t(\cdot \mid x, y, I; \phi)}
\left[
\sum_{i=1}^{m}
\operatorname{Div}_\text{KL}\left(
\pi_s(\cdot \mid x, \tilde{y}_{<i}; \theta)
\;\|\;
\pi_t(\cdot \mid x, \tilde{y}_{<i}; \phi)
\right)
\right],
\end{equation}
where $\operatorname{Div}_{\text{KL}}(\cdot\|\cdot)$ denotes the KL divergence. \textbf{In this way, the student can benefit from the minimal distributional shift of trajectories derived from its own policy, while also leveraging the teacher’s policy to provide concise reasoning supervision.} We further provide an algorithm of MPD in Alg.~\ref{alg:mpd}.

\begin{algorithm}[htbp]
\caption{Mixed-Policy Distillation}
\label{alg:mpd}
\begin{algorithmic}[1]
\Require Teacher model $\pi_t(\cdot;\phi)$, student model $\pi_s(\cdot;\theta)$, input dataset $D$
\Require Compression instruction $I$, learning rate $\eta$, batch size $B$, number of training steps $K$
\For{each training step $k = 1,\ldots,K$}
    \State Sample a batch of questions $\{x_b\}_{b=1}^{B} \sim D$
    \For{each question $x_b$}
        \State Sample an on-policy reasoning trajectory from the student: $y_b \sim \pi_s(\cdot \mid x_b; \theta)$
        \State Compress the student trajectory with the teacher: $\tilde{y}_b \sim \pi_t(\cdot \mid x_b, y_b, I; \phi)$
    \EndFor
    \State Update $\theta$ by minimizing the MPD objective:
    \[
    \theta \leftarrow \theta - \eta \nabla_{\theta}
    \frac{1}{B}
    \sum_{b=1}^{B}
    \sum_{i=1}^{|\tilde{y}_b|}
    \operatorname{Div}_{\mathrm{KL}}
    \left(
    \pi_s(\cdot \mid x_b,\tilde{y}_{b,<i};\theta)
    \;\middle\|\;
    \pi_t(\cdot \mid x_b,\tilde{y}_{b,<i};\phi)
    \right)
    \]
\EndFor
\end{algorithmic}
\end{algorithm}
\vspace{-10pt}

\section{Experiments}
\subsection{Experimental Setup}
\label{sec:experiment_setup}
\paragraph{Datasets \& Evaluation Metrics.}
We use OpenMathInstruct-2~\footnote{\url{https://huggingface.co/datasets/nvidia/OpenMathInstruct-2}}, a massive, open-source mathematical reasoning dataset, and adopt its train\_1M split as the training data. For evaluation, following~\citep{litecot}, we use three mathematical reasoning benchmarks: GSM8K~\citep{cobbe2021training}, AIME24~\citep{aime24}, and AIME25~\citep{aime25}. To evaluate generalization beyond math, we additionally use CodeMMLU~\citep{nguyen2025codemmlu} for code reasoning and LongBench-v2~\citep{bai2025longbench} for long-context reasoning. We report Pass@4 (P@4) to measure reasoning performance, defined as whether at least one of four sampled generations answers the question correctly, and the average number of generated tokens (\#Tok) to measure compression extent.

\paragraph{MPD Implementation Details.}
We use Qwen3-8B~\citep{yang2025qwen3} as the teacher model and Qwen3-1.7B as the student model. During training, the student first generates a reasoning trajectory and final answer with a maximum length of 27,000 tokens. The teacher then compresses this trajectory into a concise reasoning trace with a maximum length of 10,000 tokens. All other generation-related hyperparameters follow the default settings of the original Qwen models. We train the student for 200 steps with a batch size of 8 on 4 NVIDIA H100 GPUs. More implementation details, including the complete hyperparameter settings, are provided in App.~\ref{appendix:training_detail}. All results are averaged over three runs with different random seeds.

\paragraph{Baselines.}
We compare MPD with five baselines. First, we consider three training-free methods: Vanilla LLM, where Qwen3-1.7B directly answers the question without explicit compression; Direct Compression (Direct Comp.), which prompts the model to reason concisely (see Table~\ref{table:dc_prompt} for the full prompt); and Chain-of-Draft\citep{DBLP:journals/corr/abs-2502-18600}, which prompts the model to keep only concise intermediate thoughts. Second, we compare with LiteCoT~\citep{litecot}, an SFT-based method that refines reasoning trajectories and fine-tunes the Qwen3-1.7B on the resulting concise traces. Third, we consider CRISP~\citep{opsdc_crisp}, an on-policy self-distillation method for concise reasoning that uses the same model as both student and teacher, with the teacher conditioned on a ``be concise’’ instruction and the student aligned on self-generated trajectories via reverse KL. Since CRISP is originally formulated as self-distillation, we adapt it to our teacher-student setting by using Qwen3-8B as the teacher and Qwen3-1.7B as the student, ensuring a fair comparison with MPD. 
All reported results are averaged over three runs with different random seeds. We provide implementation details of baseline methods in App. \ref{appendix:implementation}.
\subsection{Comparative Results}
\label{sec:results}

\paragraph{Main Results.}
As shown in Table~\ref{table:overall_color}, we have several key findings: (1) MPD outperforms almost all baselines across diverse reasoning tasks\footnote{MPD does not achieve the best overall rank on GSM8K because GSM8K is a relatively easy benchmark, where performance is already saturated even for the Vanilla LLM baseline.}. Compared with Vanilla LLM, MPD achieves a 4.5\% relative improvement in Pass@4 and a 25.6\% relative reduction in token usage on average, showing that reasoning compression can improve efficiency without sacrificing performance. We attribute this to concise reasoning reducing unnecessary iterative reflection, making the model less likely to overthink and allowing it to follow a more direct reasoning path. (2) Training-free compression methods can reduce token usage but often hurt reasoning quality. Chain-of-Draft substantially lowers token usage, but suffers from severe performance degradation, even underperforming Direct Comp., suggesting that forcing the model to produce sparse, ``draft-like’’ intermediate thoughts can disrupt the continuous logical reasoning needed for problem solving~\citep{xu2026overconfident}. (3) Fine-tuning on pre-refined traces provides only limited gains. Although LiteCoT performs slightly better than Chain-of-Draft, it still suffers from distribution mismatch caused by static off-policy supervision, since the student is not trained to compress its own reasoning trajectories. (4) On-policy distillation is helpful but insufficient without teacher-guided compression. Our adapted CRISP achieves a competitive Pass@4, benefiting from on-policy training over student-generated trajectories. However, it still trails MPD in both performance and compression because its compression signal mainly comes from brevity prompting rather than teacher-compressed student trajectories. In contrast, MPD distills over teacher-compressed versions of the student’s own reasoning traces, combining the inference-time alignment of on-policy learning with the compression advantage of the stronger teacher.



\begin{table*}[t]
\centering
\caption{Overall performance and token usage. Colors represent the relative improvement (\colorbox{DeepGreen!60!MyGreen}{Green}) or regression (\colorbox{DeepRed!60!MyRed}{Red}) compared to the Vanilla LLM baseline.}
\label{table:overall_color}
\resizebox{\columnwidth}{!}{
\setlength{\tabcolsep}{3.8pt}
\begin{NiceTabular}{l rr rr rr rr rr rr}[colortbl-like]
    \toprule
    \multirow{2}{*}{\textbf{Method}} & \multicolumn{2}{c}{\textbf{GSM8K}} & \multicolumn{2}{c}{\textbf{AIME24}} & \multicolumn{2}{c}{\textbf{AIME25}} & \multicolumn{2}{c}{\textbf{CodeMMLU}} & \multicolumn{2}{c}{\textbf{LongBench2}} & \multicolumn{2}{c}{\textbf{Avg.}} \\
    \cmidrule(lr){2-3} \cmidrule(lr){4-5} \cmidrule(lr){6-7} \cmidrule(lr){8-9} \cmidrule(lr){10-11} \cmidrule(lr){12-13}
    & \textbf{P@4} & \textbf{\#Tok} & \textbf{P@4} & \textbf{\#Tok} & \textbf{P@4} & \textbf{\#Tok} & \textbf{P@4} & \textbf{\#Tok} & \textbf{P@4} & \textbf{\#Tok} & \textbf{P@4} & \textbf{\#Tok} \\ 
    \midrule
    Vanilla LLM & 94.5 & 2.1k & 63.3 & 16.8k & 53.3 & 17.7k & 81.2 & 3.6k & 48.7 & 0.9k & 68.2 & 8.2k \\
    \midrule
    
    Direct Comp. & 
    \cellcolor{DeepRed!10!MyRed} 94.1 & \cellcolor{DeepGreen!30!MyGreen} 1.2k & 
    \cellcolor{DeepGreen!30!MyGreen} 70.0 & \cellcolor{DeepGreen!20!MyGreen} 14.5k & 
    \cellcolor{DeepRed!15!MyRed} 50.0 & \cellcolor{DeepGreen!20!MyGreen} 15.2k & 
    \cellcolor{DeepGreen!20!MyGreen} 82.0 & \cellcolor{DeepGreen!40!MyGreen} 2.4k & 
    \cellcolor{DeepRed!10!MyRed} 48.1 & \cellcolor{DeepGreen!25!MyGreen} 0.8k & 
    \cellcolor{DeepGreen!15!MyGreen} 68.9 & \cellcolor{DeepGreen!30!MyGreen} 6.9k \\
    
    Chain-of-Draft & 
    \cellcolor{DeepRed!85!MyRed} 42.8 & \cellcolor{DeepGreen!80!MyGreen} 0.6k & 
    \cellcolor{white} 63.3 & \cellcolor{DeepGreen!65!MyGreen} 11.8k & 
    \cellcolor{DeepRed!15!MyRed} 50.0 & \cellcolor{DeepGreen!45!MyGreen} 13.0k & 
    \cellcolor{DeepRed!30!MyRed} 79.0 & \cellcolor{DeepGreen!70!MyGreen} 2.0k & 
    \cellcolor{DeepRed!45!MyRed} 40.8 & \cellcolor{DeepRed!60!MyRed} 1.8k & 
    \cellcolor{DeepRed!70!MyRed} 55.1 & \cellcolor{DeepGreen!65!MyGreen} 5.9k \\
    
    LiteCoT & 
    \cellcolor{DeepRed!75!MyRed} 52.0 & \cellcolor{DeepGreen!30!MyGreen} 1.2k & 
    \cellcolor{DeepGreen!20!MyGreen} 66.7 & \cellcolor{DeepGreen!10!MyGreen} 15.6k & 
    \cellcolor{DeepRed!30!MyRed} 46.7 & \cellcolor{DeepGreen!30!MyGreen} 14.6k & 
    \cellcolor{DeepGreen!15!MyGreen} 82.1 & \cellcolor{DeepGreen!10!MyGreen} 3.4k & 
    \cellcolor{DeepRed!65!MyRed} 36.6 & \cellcolor{DeepRed!75!MyRed} 2.0k & 
    \cellcolor{DeepRed!55!MyRed} 56.8 & \cellcolor{DeepGreen!20!MyGreen} 7.3k \\
    
    CRISP & 
    \cellcolor{DeepRed!5!MyRed} 94.3 & \cellcolor{DeepGreen!25!MyGreen} 1.7k & 
    \cellcolor{DeepGreen!30!MyGreen} 70.0 & \cellcolor{DeepGreen!20!MyGreen} 14.8k & 
    \cellcolor{DeepGreen!30!MyGreen} 56.6 & \cellcolor{DeepGreen!15!MyGreen} 16.3k & 
    \cellcolor{DeepGreen!25!MyGreen} 82.7 & \cellcolor{DeepRed!20!MyRed} 4.1k & 
    \cellcolor{DeepGreen!25!MyGreen} 49.9 & \cellcolor{DeepRed!20!MyRed} 1.0k & 
    \cellcolor{DeepGreen!25!MyGreen} 70.7 & \cellcolor{DeepGreen!20!MyGreen} 7.6k \\
    
    \textbf{MPD (Ours)} & 
    \cellcolor{DeepRed!5!MyRed} 94.2 & \cellcolor{DeepGreen!50!MyGreen} 1.4k & 
    \cellcolor{DeepGreen!80!MyGreen} 73.3 & \cellcolor{DeepGreen!55!MyGreen} 13.4k & 
    \cellcolor{DeepGreen!80!MyGreen} 56.7 & \cellcolor{DeepGreen!75!MyGreen} 12.9k & 
    \cellcolor{DeepGreen!60!MyGreen} 82.8 & \cellcolor{DeepGreen!60!MyGreen} 2.3k & 
    \cellcolor{DeepGreen!55!MyGreen} 50.3 & \cellcolor{DeepGreen!80!MyGreen} 0.7k & 
    \cellcolor{DeepGreen!70!MyGreen} 71.3 & \cellcolor{DeepGreen!65!MyGreen} 6.1k \\

    \midrule
    \textbf{MPD Rank} &
    3rd & 4th &
    1st & 2nd &
    1st & 1st &
    1st & 2nd &
    1st & 1st &
    1st & 2nd \\
    
    \bottomrule
\end{NiceTabular}}
\end{table*}
\begin{table*}[t]
\centering
\caption{Ablation study on different policy variants. Colors represent the relative improvement (\colorbox{DeepGreen!60!MyGreen}{Green}) or regression (\colorbox{DeepRed!60!MyRed}{Red}) compared to the Vanilla LLM baseline.}
\label{table:ablation_color}
\resizebox{\columnwidth}{!}{
\setlength{\tabcolsep}{3.8pt}
\begin{NiceTabular}{l rr rr rr rr rr rr}[colortbl-like]
    \toprule
    \multirow{2}{*}{\textbf{Method}} & \multicolumn{2}{c}{\textbf{GSM8K}} & \multicolumn{2}{c}{\textbf{AIME24}} & \multicolumn{2}{c}{\textbf{AIME25}} & \multicolumn{2}{c}{\textbf{CodeMMLU}} & \multicolumn{2}{c}{\textbf{LongBench2}} & \multicolumn{2}{c}{\textbf{Avg.}} \\
    \cmidrule(lr){2-3} \cmidrule(lr){4-5} \cmidrule(lr){6-7} \cmidrule(lr){8-9} \cmidrule(lr){10-11} \cmidrule(lr){12-13}
    & \textbf{P@4} & \textbf{\#Tok} & \textbf{P@4} & \textbf{\#Tok} & \textbf{P@4} & \textbf{\#Tok} & \textbf{P@4} & \textbf{\#Tok} & \textbf{P@4} & \textbf{\#Tok} & \textbf{P@4} & \textbf{\#Tok} \\ 
    \midrule
    
    Off-Policy & 
    \cellcolor{DeepRed!10!MyRed} 94.2 & \cellcolor{DeepRed!35!MyRed} 2.8k & 
    \cellcolor{DeepGreen!35!MyGreen} 70.0 & \cellcolor{DeepRed!50!MyRed} 22.9k & 
    \cellcolor{white} 53.3 & \cellcolor{DeepRed!40!MyRed} 21.9k & 
    \cellcolor{DeepGreen!25!MyGreen} 82.4 & \cellcolor{DeepRed!60!MyRed} 5.4k & 
    \cellcolor{DeepGreen!30!MyGreen} 51.7 & \cellcolor{DeepRed!20!MyRed} 1.0k & 
    \cellcolor{DeepGreen!25!MyGreen} 70.3 & \cellcolor{DeepRed!45!MyRed} 10.6k \\
    
    Off-Policy (C) & 
    \cellcolor{DeepGreen!5!MyGreen} 94.6 & \cellcolor{DeepRed!25!MyRed} 2.7k & 
    \cellcolor{DeepGreen!20!MyGreen} 66.6 & \cellcolor{DeepRed!35!MyRed} 20.8k & 
    \cellcolor{DeepGreen!30!MyGreen} 56.6 & \cellcolor{DeepRed!30!MyRed} 20.5k & 
    \cellcolor{DeepGreen!15!MyGreen} 82.2 & \cellcolor{DeepRed!35!MyRed} 4.5k & 
    \cellcolor{DeepGreen!10!MyGreen} 49.0 & \cellcolor{white} 0.9k & 
    \cellcolor{DeepGreen!15!MyGreen} 69.8 & \cellcolor{DeepRed!30!MyRed} 9.9k \\
    
    On-Policy & 
    \cellcolor{DeepRed!15!MyRed} 94.0 & \cellcolor{DeepRed!60!MyRed} 3.1k & 
    \cellcolor{DeepGreen!80!MyGreen} 73.3 & \cellcolor{DeepRed!40!MyRed} 20.9k & 
    \cellcolor{white} 53.3 & \cellcolor{DeepRed!40!MyRed} 21.9k & 
    \cellcolor{DeepGreen!70!MyGreen} 84.2 & \cellcolor{DeepRed!60!MyRed} 5.2k & 
    \cellcolor{DeepRed!20!MyRed} 47.9 & \cellcolor{DeepRed!20!MyRed} 1.0k & 
    \cellcolor{DeepGreen!35!MyGreen} 70.5 & \cellcolor{DeepRed!70!MyRed} 11.8k \\
    
    On-Policy (C) & 
    \cellcolor{DeepRed!15!MyRed} 94.1 & \cellcolor{white} 2.1k & 
    \cellcolor{DeepGreen!60!MyGreen} 70.0 & \cellcolor{DeepGreen!15!MyGreen} 15.8k & 
    \cellcolor{DeepGreen!65!MyGreen} 56.7 & \cellcolor{DeepGreen!15!MyGreen} 17.0k & 
    \cellcolor{DeepGreen!45!MyGreen} 82.8 & \cellcolor{white} 3.6k & 
    \cellcolor{DeepRed!10!MyRed} 48.3 & \cellcolor{DeepRed!20!MyRed} 1.0k & 
    \cellcolor{DeepGreen!40!MyGreen} 70.5 & \cellcolor{DeepGreen!10!MyGreen} 7.9k \\
    
    \textbf{MPD (Ours)} & 
    \cellcolor{DeepRed!10!MyRed} 94.2 & \cellcolor{DeepGreen!60!MyGreen} 1.4k & 
    \cellcolor{DeepGreen!80!MyGreen} 73.3 & \cellcolor{DeepGreen!60!MyGreen} 13.4k & 
    \cellcolor{DeepGreen!80!MyGreen} 56.7 & \cellcolor{DeepGreen!75!MyGreen} 12.9k & 
    \cellcolor{DeepGreen!45!MyGreen} 82.8 & \cellcolor{DeepGreen!70!MyGreen} 2.3k & 
    \cellcolor{DeepGreen!45!MyGreen} 50.3 & \cellcolor{DeepGreen!80!MyGreen} 0.7k & 
    \cellcolor{DeepGreen!75!MyGreen} 71.3 & \cellcolor{DeepGreen!75!MyGreen} 6.1k \\
    \midrule
    \textbf{MPD Rank} &
    2rd & 1st &
    1st & 1st &
    1st & 1st &
    2nd & 1st &
    2nd & 1st &
    1st & 1st \\
    \bottomrule
\end{NiceTabular}}
\vspace{-10pt}
\end{table*}
\paragraph{Ablation Studies.}
To better demonstrate MPD’s advantage, we perform ablation studies and report results in Table~\ref{table:ablation_color} on four controlled variants: Off-Policy denotes standard off-policy distillation, where the Qwen3-8B teacher generates full reasoning trajectories and the Qwen3-1.7B student is trained to match the teacher on these teacher-generated traces. On-Policy denotes standard on-policy distillation, where the Qwen3-1.7B student generates its own reasoning trajectories and is then trained to match the Qwen3-8B teacher under these student-generated traces. We also consider compressed variants using the same compression prompt as MPD. Off-Policy (C) denotes compressed off-policy distillation, where the teacher both generates and compresses the trajectories before the student is trained on the compressed teacher traces. On-Policy (C) denotes compressed on-policy distillation, where the student generates and compresses its own trajectories and is then trained to match the teacher on the resulting compressed traces. We have several findings: (1) On-Policy achieves good performance by training on the student’s own trajectory distribution, but it also produces the highest token usage. This suggests that on-policy learning reduces distribution mismatch, but directly learning from verbose student trajectories can reinforce redundant reasoning patterns. (2) Off-Policy uses shorter teacher-generated trajectories, but does not outperform On-Policy, confirming the mismatch between teacher trajectories and the student’s own inference-time behaviour limits effective learning. (3) Explicit compression using MPD's compression prompt consistently reduces token usage compared with the corresponding verbose variants. However, On-Policy (C) relies on the weaker student to compress its own trajectories, which can potentially remove useful reasoning steps and weaken performance, while Off-Policy (C) benefits from teacher compression but still suffers from off-policy supervision. (4) MPD achieves the best trade-off between performance and conciseness by combining the two necessary components: student-generated trajectories provide the on-policy reasoning backbone, and teacher-guided compression removes redundancy while preserving the underlying reasoning structure.

\paragraph{Why MPD Outperforms: a Perplexity Perspective.}

\begin{figure}[t]
    \centering
    \begin{subfigure}{0.48\linewidth}
        \centering
            \includegraphics[width=\linewidth]{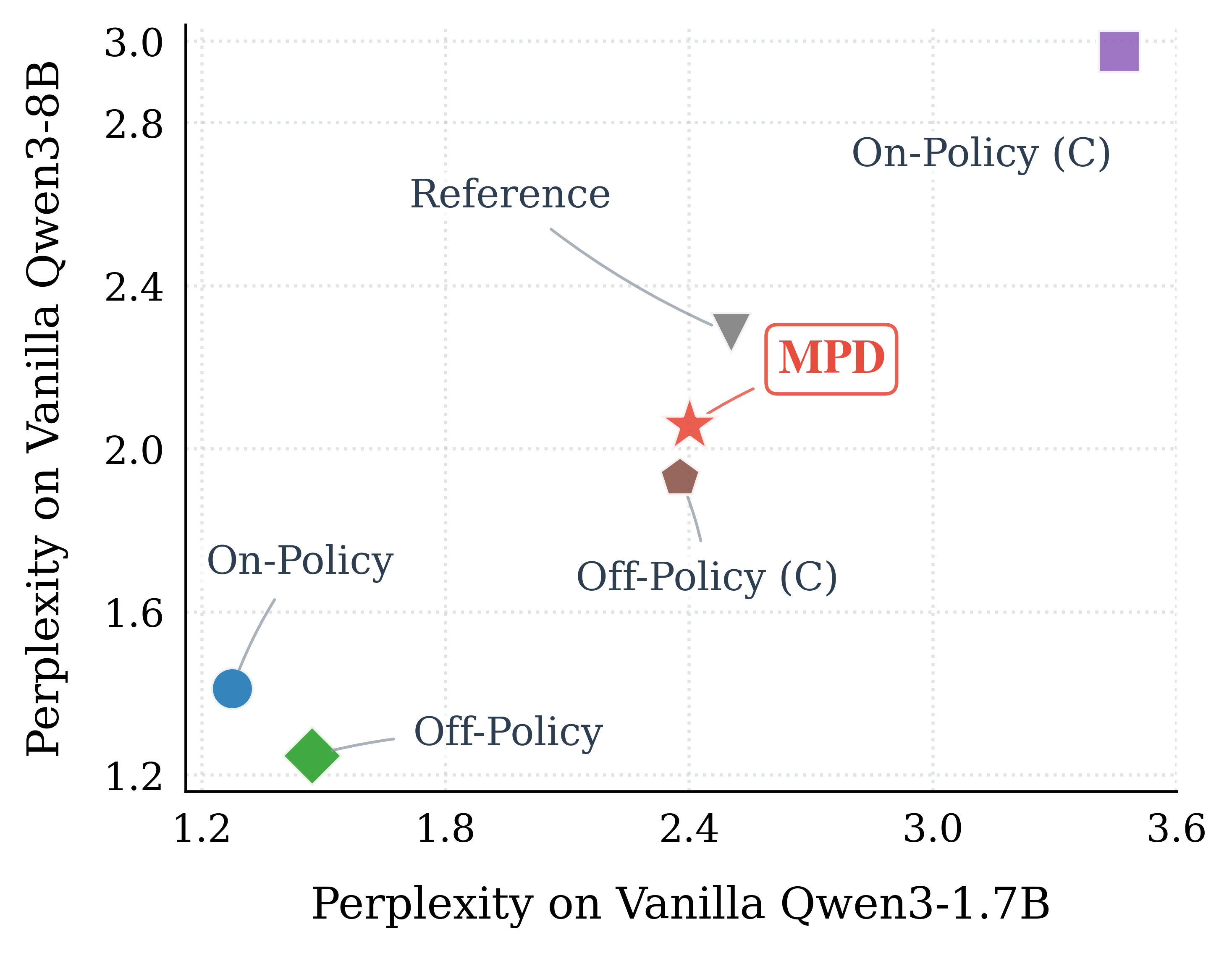}
        \caption{Comparison among different training methods.}
        \label{fig:ppl1}
    \end{subfigure}
    \hfill
    \begin{subfigure}{0.48\linewidth}
        \centering
        \includegraphics[width=\linewidth]{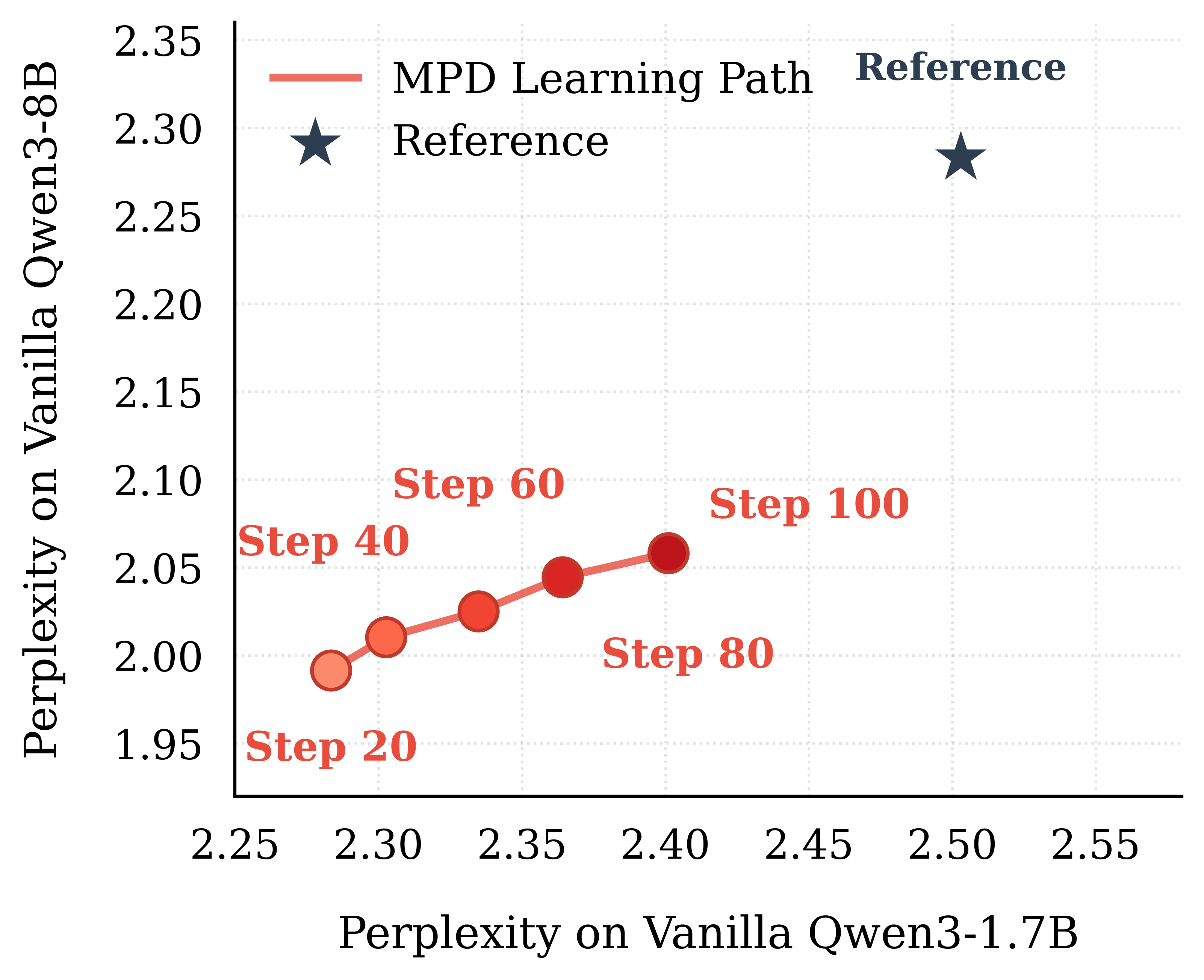}
        \caption{MPD learning path.}
        \label{fig:ppl2}
    \end{subfigure}
    \caption{Perplexity analysis of generated reasoning traces from different training methods under vanilla Qwen models.}\label{fig:ppl}
    \vspace{-10pt}
\end{figure}

To better understand the effect of MPD, we analyze the perplexity of generated reasoning trajectories under the vanilla Qwen3-1.7B and Qwen3-8B. Specifically, we sample the same 100 questions from OpenMathInstruct-2 and let different trained models generate their own reasoning traces. We then evaluate each set of traces by computing the average perplexity across all questions under vanilla models. We also include the human-inspected reference solutions from OpenMathInstruct-2, which are expected to be non-redundant.
As shown in Fig.~\ref{fig:ppl1}, the reference solutions do not have the lowest perplexity under either vanilla model, suggesting that concise, human-inspected reasoning is not simply the most likely continuation under the original model distributions. Standard On-Policy and Off-Policy trajectories have much lower perplexity under both vanilla models, indicating that they remain close to typical model-generated reasoning, but they are also far from the reference region. After explicit compression, both On-Policy (C) and Off-Policy (C) move toward the reference, showing that compression helps shift generated traces toward more concise reasoning. However, they do so differently. On-Policy (C) has high perplexity under both vanilla models, likely because the weaker student both generates and compresses its own traces. This can produce compressed trajectories that deviate from ordinary student generations while still lacking the teacher’s stronger ability to organize concise reasoning, making them less natural to both vanilla models. Off-Policy (C), in contrast, stays closer to vanilla Qwen3-8B than On-Policy (C), since both generation and compression are performed by the teacher. However, since its trajectories are generated entirely from the teacher policy, they remain less aligned with the vanilla Qwen3-1.7B distribution. MPD is closest to the reference among all variants: it uses student-generated traces to preserve the student-side distributional signal, while applying teacher compression to move the traces toward concise, reference-like reasoning.
Finally, Fig.~\ref{fig:ppl2} shows the MPD learning path across training checkpoints. As training progresses from Step 20 to Step 100, the generated trajectories gradually move toward the reference solutions. This indicates that MPD provides a consistent learning signal that shifts the student away from redundant reasoning traces and toward more concise, reference-like reasoning behaviour.

\paragraph{Case Study.} 
\begin{table*}[t]
\centering
\small
\caption{Case study comparing MPD with On-Policy, On-Policy (C), and Off-Policy (C). Text in \textbf{\textcolor[rgb]{0.7,0.3,0.3}{red}} marks inefficient or incorrect reasoning, while text in \textbf{\textcolor[rgb]{0.13,0.55,0.8}{blue}} marks MPD's key correct reasoning.}
\renewcommand\arraystretch{1.18}
\setlength{\fboxsep}{1.0pt}
\resizebox{\columnwidth}{!}{
\begin{tabular}{p{0.88\textwidth}}
\hline
\rowcolor{gray!8}{\textbf{Case:} MPD vs. On-Policy, On-Policy (C), and Off-Policy (C).}\\
\hline
\textbf{Question:} Let \(A\), \(B\), \(C\), and \(D\) be points on the hyperbola
\(\frac{x^2}{20}-\frac{y^2}{24}=1\) such that \(ABCD\) is a rhombus whose diagonals intersect at the origin. Find the greatest real number that is less than \(BD^2\) for all such rhombi.\\
\textbf{Ground-Truth:} \(\boxed{480}\)\\
\textbf{On-Policy Response (13.3k tokens, correct but verbose):}
Okay, so I need to solve this problem about a rhombus inscribed in a hyperbola. \(\ldots\)
\textbf{\textcolor[rgb]{0.7,0.3,0.3}{Wait, so \(BD^2\) can take various values, and we need to find the greatest number that is less than \(BD^2\) for all such rhombi. That is, the infimum of \(BD^2\)? Wait, no.}} \(\ldots\)
After a long derivation, OPD obtains \(BD^2 = 480\cdot \frac{t}{t-\frac{600}{11}}\).
Now, as \(t\to\infty\), the expression approaches \(480\). Thus, the infimum of \(BD^2\) is \(480\), and since \(BD^2>480\) for all such rhombi, the greatest real number less than \(BD^2\) for all such rhombi is \(\boxed{480}\).\\

\textbf{On-Policy (C) Response (7.7k tokens, incorrect):}
Okay, so I need to solve this problem about a rhombus inscribed in the hyperbola \(x^2/20-y^2/24=1\). \(\ldots\)
\textbf{\textcolor[rgb]{0.7,0.3,0.3}{Therefore, \(BD^2 = 4(r^2+s^2)=80+\frac{22}{3}s^2\).}}
\textbf{\textcolor[rgb]{0.7,0.3,0.3}{So, since \(s^2\) is non-negative, \(BD^2\) is at least \(80\). Therefore, the minimal \(BD^2\) is \(80\).}} \(\ldots\)
\textbf{\textcolor[rgb]{0.7,0.3,0.3}{Therefore, boxed answer is \(\boxed{80}\).}}\\

\textbf{Off-Policy (C) Response (22.2k tokens, incorrect):}
Okay, so I need to solve this problem about a rhombus inscribed in a hyperbola. \(\ldots\)
\textbf{\textcolor[rgb]{0.7,0.3,0.3}{This is very confusing. Given that the problem states such a rhombus exists, I must have made a mistake in my reasoning.}} \(\ldots\)
\textbf{\textcolor[rgb]{0.7,0.3,0.3}{Therefore, \(BD^2 = 4/k\), where \(k \in [-1/24,1/20]\). Since \(k\) is in this range, and \(BD^2=4/k\), when \(k\) is negative, \(BD^2\) is negative, which is impossible.}} \(\ldots\)
\textbf{\textcolor[rgb]{0.7,0.3,0.3}{Therefore, the answer is \(80\), which is \(\boxed{80}\).}}\\

\textbf{MPD Response (6.2k tokens, correct):}
Okay, let's try to tackle this problem. \(\ldots\)
\textbf{\textcolor[rgb]{0.13,0.55,0.8}{\(BD^2\) is \(16(5\sec^2\phi+6\tan^2\phi)\).}}
The diagonals are perpendicular, giving \(\sin\theta\sin\phi=-5/6\). \(\ldots\)
\textbf{\textcolor[rgb]{0.13,0.55,0.8}{Therefore, \(BD^2=16(5+11\tan^2\phi)\).}}
\textbf{\textcolor[rgb]{0.13,0.55,0.8}{If \(\tan^2\phi\geq 25/11\), then \(5+11\tan^2\phi\geq 30\), so \(BD^2\geq 16\cdot 30=480\).}}
\textbf{\textcolor[rgb]{0.13,0.55,0.8}{Therefore, the answer is \(\boxed{480}\).}}\\
\hline
\end{tabular}}
\label{tab:case-study-mpd-baselines}
\vspace{-10pt}
\end{table*}
We present a case study on an AIME24 question comparing MPD with On-Policy, On-Policy (C), and Off-Policy (C), with more examples in App.\ref{appendix:case_study}. As shown in Table~\ref{tab:case-study-mpd-baselines}, MPD is the only method that is both correct and concise. Compared with On-Policy, which eventually reaches the correct answer but spends 13.3k tokens with repeated verification and unnecessary detours, MPD obtains the same answer using only 6.2k tokens, reducing the reasoning length by 53.38\%. On-Policy (C) shortens the response to 7.7k tokens, but drops a crucial geometric constraint and incorrectly concludes that the answer is \(80\). Off-Policy (C) is even longer, using 22.2k tokens, and still fails by following a confusing parametrization that leads to the same wrong answer. These failures show that compression without reliable reasoning can remove essential structure, while teacher-generated compressed traces alone may not align well with the student’s own problem-solving process. MPD combines student-side exploration with teacher-guided compression, thus achieving concise reasoning without discarding the key intermediate structure needed for correctness.
\vspace{-5pt}
\section{Conclusion}
We propose Mixed-Policy Distillation (MPD), a reasoning compression framework that transfers concise reasoning behaviour from a larger teacher model to a smaller student model without requiring ground-truth answers, reward models, length penalties, or externally curated compressed traces. Motivated by the observation that larger reasoning models often solve the same problems with shorter traces, MPD samples trajectories from the student and uses the larger teacher to rewrite them into more concise reasoning traces. The student is then trained on these teacher-compressed student trajectories through KL-based distillation. Experiments across mathematical reasoning, code reasoning, and long-context reasoning benchmarks show that MPD consistently reduces token usage while maintaining or improving performance. Further analysis demonstrates that MPD’s advantage comes from combining student-side exploration with teacher-guided compression, enabling concise reasoning without discarding the key intermediate structure needed for correctness. Overall, MPD provides a practical path toward efficient small-model reasoning, reducing latency and computational cost while preserving strong problem-solving ability. 


\section{Limitations \& Future Directions}
\label{limitation_main}
MPD provides an efficient learning signal by aligning training with the student’s reasoning patterns while making trajectories concise. However, it does not explicitly optimize for correctness, so incorrect student-teacher trajectories may still be learned; adding correctness-aware signals could improve performance. MPD also requires full trajectories for teacher compression, which may be difficult under limited context windows. Future work can explore multi-step compression, where the teacher iteratively refines partial trajectories. See extended discussion in App.~\ref{limitation}.

\paragraph{Acknowledgment.} We sincerely thank Prof. Andreas Vlachos for his valuable feedback and continuous support of Zifeng throughout this work. 
Han Yang received funding from the Deutsche Forschungsgemeinschaft (DFG) under grant number: MA 3964/15-3 (SocioHub project). Han Yang received additional funding from the European Union under the Horizon Europe grant OMINO -- Overcoming Multilevel INformation Overload\footnote{\url{https://ominoproject.eu/}} under grant number 101086321 \cite{informationoverload}.

\small
\bibliography{section/references}
\bibliographystyle{unsrt}

\clearpage

\appendix

\section{Additional Experimental Detail}
\label{appendix:training_detail}

\subsection{Experimental Configurations}
\begin{figure}[t]
  \centering
    \centering
    \includegraphics[width=0.5\textwidth]{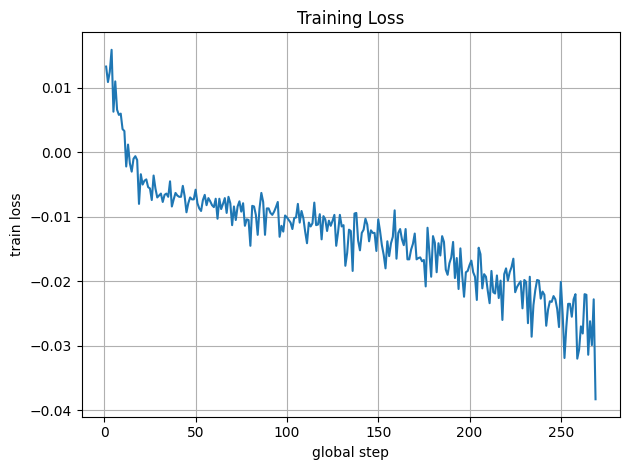}
    \caption{Training loss.}
    \label{fig:train_loss}
\end{figure}

\begin{table}[t]
\centering
\caption{Hyper-parameter configurations for MPD.}
\label{table:configs}
\small
\resizebox{\columnwidth}{!}{
\begin{tabular}{llll}
\toprule
\multicolumn{2}{c}{\textbf{Training \& Loss}} & \multicolumn{2}{c}{\textbf{Sampling (Distillation)}} \\
\midrule
Dataset                 & OpenMathInstruct-2     & Max Completion (Student) & 27,000 \\
Student Model           & Qwen3-1.7B             & Max Completion (Teacher) & 10,000 \\
Teacher Model           & Qwen3-8B               & Max Total Length         & 30,000 \\
Training Steps          & 200                    & Temperature              & 0.6    \\
Epochs                  & 1                      & Top-p                    & 0.95   \\
Batch Size (per device) & 2                      & Top-k                    & 20     \\
Learning Rate           & 5.00E-06               & \multicolumn{2}{c}{\textbf{Evaluation}} \\
\cmidrule{3-4}
Beta ($\beta$)          & 1                      & Max Total Length         & 32,768 \\
Lambda ($\lambda$)      & 1                      & Temperature              & 0.6    \\
JSD Token Clip          & 0.05                   & Top-p                    & 0.95   \\
Max Grad Norm           & 0.1                    & Top-k                    & 20     \\
\midrule
\multicolumn{4}{c}{\textbf{LoRA Configuration}} \\
\midrule
Rank ($r$)              & 64                     & Alpha ($\alpha$)         & 128    \\
\bottomrule
\end{tabular}}
\end{table}
\paragraph{Experimental Configurations.} We provide our training and evaluation configurations for our method in Table \ref{table:configs}. 
\paragraph{Training Loss.} We show the loss during the training in Fig.~\ref{fig:train_loss}.

\paragraph{Prompts.}
\begin{table}[t]
\centering
\caption{Prompt for the student model in answering questions.}
\label{table:prompt_student}
\small
\begin{tabularx}{\columnwidth}{|X|}
\hline
\rowcolor[gray]{0.9} \textbf{Prompt for the Student} \\ \hline
Problem: \{$x$\} \\ 
\\
Please reason step by step, and put your final answer within \textbackslash boxed\{\}. \\ \hline
\end{tabularx}
\end{table}
\begin{table}[ht]
\centering
\caption{Prompt for Direct Comp..}
\label{table:dc_prompt}
\small
\begin{tabularx}{\columnwidth}{|X|}
\hline
\rowcolor[gray]{0.9} \textbf{Prompt for Direct Comp.} \\ \hline
Solve the following problem. Give a full solution, but the final answer MUST be written in the form: \textbackslash boxed\{\}. \\
Solve the following math problem concisely and correctly. Be direct -- avoid unnecessary elaboration, redundant steps, or restating the problem. Focus only on the key reasoning steps needed to reach the answer. \\
\\
Problem: \{$x$\} \\ \hline
\end{tabularx}
\end{table}
We show here the prompts that we used during the training. The student prompt is shown in Table~\ref{table:prompt_student}. The teacher compression prompt is shown in Table~\ref{table:prompt_teacher}. The prompt used in Direct Comp. is shown in Table~\ref{table:dc_prompt}

\subsection{Implementation Detail}
\label{appendix:implementation}

\paragraph{Hardware and Software.}
All experiments involving LLMs are implemented with PyTorch~\cite{pytorch}, HuggingFace Transformers~\cite{DBLP:journals/corr/abs-1910-03771}, and vLLM~\cite{kwon2025vllm}. Model training is conducted using the TRL\footnote{\url{https://github.com/huggingface/trl}} library on a SLURM-managed computing cluster. Each training node is equipped with 96 CPU cores and 4 NVIDIA H100 GPUs, each with 94GB of memory.
For inference, we use vLLM for efficient batched generation. The decoding hyperparameters, including temperature and top-$p$, follow the recommended settings of the corresponding models, with thinking mode enabled~\cite{yang2025qwen3}.

\paragraph{Training Details.}
Our training framework is implemented on top of the OPSD repository~\cite{opsd}.
The training process took approximately 18 hours in wall-clock time, corresponding to a total computational cost of roughly 72 GPU-hours.
For baseline methods, we follow the official implementations and default hyperparameter settings whenever available. Due to software and hardware compatibility issues, we were unable to successfully execute the original implementation of CRISP~\cite{opsdc_crisp}. As an alternative, we re-implemented it based on the OPSD codebase to ensure comparable evaluation. For LiteCoT, we use the LiteCoT training data~\footnote{\url{https://huggingface.co/datasets/SmallDoge/SmallThoughts}} to train Qwen3-1.7B using the official released code on 1 NVIDIA H100 GPU.

\paragraph{Hyperparameter Search.}
The maximum completion length is determined by balancing three factors: (1) ensuring that the student model can generate the complete trajectory, (2) reserving sufficient context space for the teacher model to produce the compressed response conditioned on the generated trajectory, and (3) avoiding out-of-memory (OOM) issues during training and inference.
The batch size is adjusted according to the sequence length to fit within GPU memory constraints and prevent OOM errors. All other hyperparameters follow the recommended or default settings of the corresponding models and libraries.





\section{Additional Case Study}\label{appendix:case_study}
Table~\ref{table:case-study-vanilla-wrong-mpd-correct} illustrates a case where the Vanilla LLM produces a long but ultimately incorrect reasoning trajectory. The Vanilla response repeatedly assumes that the valid sets \(B\) behave like subsets of \(A\), leading to the erroneous equation \(2^m-1=2024\). Although the model later recognizes that this interpretation may be problematic, its self-reflection does not recover the correct counting principle. Instead, it continues to reason around the contradiction, eventually hypothesizing a typo in the problem and returning the wrong answer \(66\). In contrast, MPD identifies the key combinatorial structure directly: for a fixed maximum \(x\in A\), there are \(2^{x-1}\) possible sets. This reduces the problem to the binary expansion of \(2024\), yielding \(A=\{4,6,7,8,9,10,11\}\) and the correct answer \(55\). This example shows that longer reasoning is not necessarily more reliable; redundant reflection can amplify an early misconception, while MPD preserves the decisive reasoning steps.

Table~\ref{table:case-study-both-correct-mpd-efficient} shows a complementary case where both methods reach the correct answer, but with substantially different reasoning efficiency. The Vanilla LLM derives \(xy=25\) early, but then spends many additional steps questioning whether the shortcut is valid, setting up a transcendental equation, and attempting numerical verification. Since the problem already states that such \(x,y>1\) exist, these extra steps are unnecessary for determining \(xy\). MPD instead introduces \(k=\log_y x\), immediately obtains \(x=10k\) and \(y=\frac{5}{2k}\), and directly concludes \(xy=25\). Thus, MPD not only reduces token usage, but also encourages more goal-directed reasoning by retaining the minimal algebraic path needed to solve the problem.

Overall, these case studies suggest that MPD improves reasoning efficiency in two ways. First, it can avoid verbose but unproductive self-reflection that fails to correct an initial error. Second, even when the baseline eventually answers correctly, MPD can remove unnecessary verification detours and reach the answer with a much shorter reasoning trajectory.

\begin{table*}[t]
\centering
\small
\caption{Case study where Vanilla LLM produces a long self-reflective but incorrect solution. Text in \textbf{\textcolor[rgb]{0.7,0.3,0.3}{red}} marks the erroneous assumption, repeated confusion, and wrong final answer; text in \textbf{\textcolor[rgb]{0.13,0.55,0.8}{blue}} marks MPD's key correct reasoning path.}
\renewcommand\arraystretch{1.18}
\setlength{\fboxsep}{1.0pt}
\resizebox{\columnwidth}{!}{
\begin{tabular}{p{0.88\textwidth}}
\hline
\rowcolor{gray!8}{\textbf{Case 1:} Vanilla LLM is long, self-reflective, and incorrect; MPD is concise and correct.}\\
\hline
\textbf{Question:} Alice chooses a set \(A\) of positive integers. Then Bob lists all finite nonempty sets \(B\) of positive integers with the property that the maximum element of \(B\) belongs to \(A\). Bob's list has 2024 sets. Find the sum of the elements of \(A\).\\
\textbf{Ground-Truth:} \(\boxed{55}\)\\
\textbf{Vanilla LLM Response (26.5k tokens):}
Okay, let's try to tackle this problem step by step. \(\ldots\)
Wait, if \(A\) is a set of positive integers, let's suppose that \(A\) has elements \(a_1, a_2, \ldots, a_m\), sorted in increasing order. Then, for each \(a_i\), the number of elements in \(A\) less than \(a_i\) is \(i - 1\). Therefore, \textbf{\textcolor[rgb]{0.7,0.3,0.3}{the total number of subsets \(B\) is \(\sum_{i=1}^m 2^{i - 1}=2^m-1\)}}. \(\ldots\)
So then, if the problem says that the number of subsets is 2024, then \textbf{\textcolor[rgb]{0.7,0.3,0.3}{\(2^m - 1 = 2024\). Therefore, \(2^m = 2025\). But 2025 is not a power of 2.}} \(\ldots\)
Oh! That's a crucial mistake. So, the problem is not about subsets of \(A\), but about all finite nonempty sets \(B\) where the maximum element of \(B\) is in \(A\). \(\ldots\)
\textbf{\textcolor[rgb]{0.7,0.3,0.3}{Wait, I'm stuck. Given that, and since the problem must have a solution, I will assume that there is a typo and the number of sets is 2047, which is \(2^{11}-1\).}} \(\ldots\)
\textbf{\textcolor[rgb]{0.7,0.3,0.3}{But this is getting too long, and I don't see how this helps.}} Given that, I think the problem must have a typo, and the correct number of sets is 2047, which is \(2^{11}-1\), and the sum of the elements of \(A\) is \(1+2+\cdots+11=66\). Therefore, the answer is 66. \(\ldots\)
\textbf{\textcolor[rgb]{0.7,0.3,0.3}{Final Answer: \(\boxed{66}\)}}\\

\textbf{MPD Response (5.2k tokens):}
Okay, let's try to figure out this problem. \(\ldots\)
\textbf{\textcolor[rgb]{0.13,0.55,0.8}{But wait, if \(x\) is the maximum, then the subset must contain \(x\) and any combination of elements less than \(x\). So, for each \(x\) in \(A\), the number of subsets where the maximum is \(x\) is \(2^{x-1}\).}}
But then, the total number of subsets is the sum over all elements in \(A\) of \(2^{x-1}\). \(\ldots\)
So, the formula seems to hold. Therefore, if \(A\) has elements \(a_1,a_2,\ldots,a_k\), then the total number of subsets is \(\sum_{i=1}^k 2^{a_i-1}=2024\). \(\ldots\)
\textbf{\textcolor[rgb]{0.13,0.55,0.8}{Well, since 2024 is a sum of distinct powers of 2, as we saw, the set \(\{b_i\}\) would be the exponents in the binary representation of 2024.}} \(\ldots\)
\textbf{\textcolor[rgb]{0.13,0.55,0.8}{If we have the exponents in the binary representation of 2024 as \(3,5,6,7,8,9,10\), then the sum is \(2^3+2^5+2^6+2^7+2^8+2^9+2^{10}\).}} \(\ldots\)
Therefore, the corresponding \(b_i\) are these exponents, and then \(a_i=b_i+1\). Therefore, the elements of \(A\) are \(4,6,7,8,9,10,11\). \(\ldots\)
\textbf{\textcolor[rgb]{0.13,0.55,0.8}{Final Answer: \(\boxed{55}\)}}\\
\hline
\end{tabular}}
\label{table:case-study-vanilla-wrong-mpd-correct}
\end{table*}
\begin{table*}[t]
\centering
\small
\caption{Case study where both methods answer correctly, but Vanilla LLM spends many extra steps verifying a product already forced by the equations. Text in \textbf{\textcolor[rgb]{0.7,0.3,0.3}{red}} marks inefficient detours; text in \textbf{\textcolor[rgb]{0.13,0.55,0.8}{blue}} marks the direct reasoning that quickly leads to the correct answer.}
\renewcommand\arraystretch{1.18}
\setlength{\fboxsep}{1.0pt}
\resizebox{\columnwidth}{!}{
\begin{tabular}{p{0.88\textwidth}}
\hline
\rowcolor{gray!8}{\textbf{Case 2:} Both methods are correct, but MPD reaches the answer much more efficiently.}\\
\hline
\textbf{Question:} There exist real numbers \(x\) and \(y\), both greater than 1, such that \(\log_x(y^x)=\log_y(x^{4y})=10\). Find \(xy\).\\
\textbf{Ground-Truth:} \(\boxed{25}\)\\
\textbf{Vanilla LLM Response (8.5k tokens):}
Okay, so I need to solve this problem where there are real numbers \(x\) and \(y\), both greater than 1. \(\ldots\)
So from equation 1, \(x=10/a\). From equation 2, \(4y=10a \Rightarrow y=(10a)/4=(5a)/2\). \(\ldots\)
\textbf{\textcolor[rgb]{0.13,0.55,0.8}{\(xy=(10/a)(5a/2)=25\).}} Wait, that's interesting! The \(a\) cancels out, so \(xy=25\)? But that seems too straightforward. Let me check that again. \(\ldots\)
\textbf{\textcolor[rgb]{0.7,0.3,0.3}{Suppose that \(xy=25\). Let me check if there exist \(x\) and \(y\) greater than 1 such that \(\log_x(y^x)=\log_y(x^{4y})=10\).}} \(\ldots\)
\textbf{\textcolor[rgb]{0.7,0.3,0.3}{This is a transcendental equation in \(a\). It might not have an algebraic solution, but maybe there is a specific value of \(a\) that satisfies this equation.}} \(\ldots\)
Wait, earlier I thought that \(xy=25\) regardless of \(a\). Let me check that again. \(\ldots\)
\textbf{\textcolor[rgb]{0.7,0.3,0.3}{But to confirm, let me try to find specific \(x\) and \(y\). Suppose that \(xy=25\). Let me pick \(x=2\), then \(y=12.5\).}} \(\ldots\)
Therefore, \(a\approx6.515\). Then \(x=10/a\approx1.534\), and \(y=5a/2\approx16.2875\). \(\ldots\)
\textbf{Final Answer: \(\boxed{25}\)}\\

\textbf{MPD Response (2.6k tokens):}
Okay, let's try to solve this problem. So, we have two logarithmic equations. \(\ldots\)
Using the power rule, that becomes \(x\log_x(y)=10\). \(\ldots\)
Again, using the power rule, that's \(4y\log_y(x)=10\). \(\ldots\)
\textbf{\textcolor[rgb]{0.13,0.55,0.8}{Now, let me denote \(\log_y(x)\) as some variable to make it easier. Let me let \(\log_y(x)=k\).}}
\textbf{\textcolor[rgb]{0.13,0.55,0.8}{Then, the first equation becomes: \(x/k=10 \Rightarrow x=10k\).}}
\textbf{\textcolor[rgb]{0.13,0.55,0.8}{The second equation is \(4y\cdot k=10 \Rightarrow y\cdot k=5/2\).}}
So, we have: from the first equation, \(x=10k\); from the second equation, \(y=(5/2)/k\). \(\ldots\)
\textbf{\textcolor[rgb]{0.13,0.55,0.8}{\(xy=10k\cdot(5/2)/k=10\cdot(5/2)=25\).}} \(\ldots\)
\textbf{\textcolor[rgb]{0.13,0.55,0.8}{Final Answer: \(\boxed{25}\)}}\\
\hline
\end{tabular}}
\label{table:case-study-both-correct-mpd-efficient}
\end{table*}

\section{Limitations \& Future Directions}
\label{limitation}
\paragraph{Incorporating Correctness Objective.} Our training approach provides a dense and efficient learning signal. 
It ensures that the training trajectories remain aligned with the student’s intrinsic reasoning patterns while being concise. 
However, the correctness of the trajectories is not explicitly incorporated into the training objective. 
As a result, the model may still learn from regions where both the student and the teacher are incorrect. Incorporating correctness-aware signals into the training process may further improve performance.

\paragraph{Multi-Turn Compression.} 
Effective compression requires the student to generate a complete trajectory, and the teacher’s rewriting process also relies on access to the full trajectory. However, in scenarios with limited context windows, the student may fail to produce the complete trajectory, which prevents our method from reaching its full potential.
In such cases, multi-step compression can help maintain the quality of the compressed trajectories by allowing the teacher to iteratively refine partial outputs, thereby improving the overall learning effectiveness.

\section{Potential Societal Risks}
\label{appendix:social_risks}

\paragraph{Positive Societal Impact.}
 Our work contributes to improving the efficiency of reasoning in large language models. By enabling concise yet effective reasoning, our approach reduces the number of tokens required during inference, which can lower computational cost and energy consumption. This improves the scalability and accessibility of reasoning models, making them more practical for deployment in resource-constrained settings.
In addition, by facilitating the transfer of reasoning capabilities to smaller models, our method can help broaden access to advanced AI systems, potentially benefiting applications in education and low-resource environments. More broadly, our work encourages research on efficient reasoning, promoting approaches that balance performance and computational efficiency.

\paragraph{Potential Societal Risks.}
Methods based on model distillation or self-improving training may introduce certain risks. First, since correctness is not always explicitly enforced in the training objective, models may learn from trajectories that are incorrect, potentially leading to confident yet erroneous outputs. Second, as these approaches rely on stronger models to guide training, any biases or systematic errors present in the teacher model may be inherited or even amplified in the student model.
These risks are broadly shared across distillation-based training paradigms. Addressing them, for example by incorporating correctness-aware signals or improved verification mechanisms, remains an important direction for future work.
\clearpage




\newpage
\section*{NeurIPS Paper Checklist}

\begin{enumerate}

\item {\bf Claims}
    \item[] Question: Do the main claims made in the abstract and introduction accurately reflect the paper's contributions and scope?
    \item[] Answer: \answerYes{} 
    \item[] Justification: Yes, the main claims made in the abstract and introduction accurately reflect our contributions and scope. 
    \item[] Guidelines:
    \begin{itemize}
        \item The answer \answerNA{} means that the abstract and introduction do not include the claims made in the paper.
        \item The abstract and/or introduction should clearly state the claims made, including the contributions made in the paper and important assumptions and limitations. A \answerNo{} or \answerNA{} answer to this question will not be perceived well by the reviewers. 
        \item The claims made should match theoretical and experimental results, and reflect how much the results can be expected to generalize to other settings. 
        \item It is fine to include aspirational goals as motivation as long as it is clear that these goals are not attained by the paper. 
    \end{itemize}

\item {\bf Limitations}
    \item[] Question: Does the paper discuss the limitations of the work performed by the authors?
    \item[] Answer: \answerYes{} 
    \item[] Justification: We discussed our limitation Appendix~\ref{limitation}, we also provide future directions for solving these limitations. 
    \item[] Guidelines:
    \begin{itemize}
        \item The answer \answerNA{} means that the paper has no limitation while the answer \answerNo{} means that the paper has limitations, but those are not discussed in the paper. 
        \item The authors are encouraged to create a separate ``Limitations'' section in their paper.
        \item The paper should point out any strong assumptions and how robust the results are to violations of these assumptions (e.g., independence assumptions, noiseless settings, model well-specification, asymptotic approximations only holding locally). The authors should reflect on how these assumptions might be violated in practice and what the implications would be.
        \item The authors should reflect on the scope of the claims made, e.g., if the approach was only tested on a few datasets or with a few runs. In general, empirical results often depend on implicit assumptions, which should be articulated.
        \item The authors should reflect on the factors that influence the performance of the approach. For example, a facial recognition algorithm may perform poorly when image resolution is low or images are taken in low lighting. Or a speech-to-text system might not be used reliably to provide closed captions for online lectures because it fails to handle technical jargon.
        \item The authors should discuss the computational efficiency of the proposed algorithms and how they scale with dataset size.
        \item If applicable, the authors should discuss possible limitations of their approach to address problems of privacy and fairness.
        \item While the authors might fear that complete honesty about limitations might be used by reviewers as grounds for rejection, a worse outcome might be that reviewers discover limitations that aren't acknowledged in the paper. The authors should use their best judgment and recognize that individual actions in favor of transparency play an important role in developing norms that preserve the integrity of the community. Reviewers will be specifically instructed to not penalize honesty concerning limitations.
    \end{itemize}

\item {\bf Theory assumptions and proofs}
    \item[] Question: For each theoretical result, does the paper provide the full set of assumptions and a complete (and correct) proof?
    \item[] Answer: \answerNA{} 
    \item[] Justification: This paper contributes an approach to train a model with the ability of efficient reasoning, no theoretical assumptions or formal proofs are provided.
    \item[] Guidelines:
    \begin{itemize}
        \item The answer \answerNA{} means that the paper does not include theoretical results. 
        \item All the theorems, formulas, and proofs in the paper should be numbered and cross-referenced.
        \item All assumptions should be clearly stated or referenced in the statement of any theorems.
        \item The proofs can either appear in the main paper or the supplemental material, but if they appear in the supplemental material, the authors are encouraged to provide a short proof sketch to provide intuition. 
        \item Inversely, any informal proof provided in the core of the paper should be complemented by formal proofs provided in appendix or supplemental material.
        \item Theorems and Lemmas that the proof relies upon should be properly referenced. 
    \end{itemize}

    \item {\bf Experimental result reproducibility}
    \item[] Question: Does the paper fully disclose all the information needed to reproduce the main experimental results of the paper to the extent that it affects the main claims and/or conclusions of the paper (regardless of whether the code and data are provided or not)?
    \item[] Answer: \answerYes{} 
    \item[] Justification: We provide detailed descriptions of the proposed method,
model configurations, prompting strategies, datasets,
evaluation scripts, and experimental settings, necessary
to reproduce the main results of this work. Details can be found in Section~\ref{sec:experiment_setup} and Appendix~\ref{appendix:training_detail}. 
    \item[] Guidelines:
    \begin{itemize}
        \item The answer \answerNA{} means that the paper does not include experiments.
        \item If the paper includes experiments, a \answerNo{} answer to this question will not be perceived well by the reviewers: Making the paper reproducible is important, regardless of whether the code and data are provided or not.
        \item If the contribution is a dataset and\slash or model, the authors should describe the steps taken to make their results reproducible or verifiable. 
        \item Depending on the contribution, reproducibility can be accomplished in various ways. For example, if the contribution is a novel architecture, describing the architecture fully might suffice, or if the contribution is a specific model and empirical evaluation, it may be necessary to either make it possible for others to replicate the model with the same dataset, or provide access to the model. In general. releasing code and data is often one good way to accomplish this, but reproducibility can also be provided via detailed instructions for how to replicate the results, access to a hosted model (e.g., in the case of a large language model), releasing of a model checkpoint, or other means that are appropriate to the research performed.
        \item While NeurIPS does not require releasing code, the conference does require all submissions to provide some reasonable avenue for reproducibility, which may depend on the nature of the contribution. For example
        \begin{enumerate}
            \item If the contribution is primarily a new algorithm, the paper should make it clear how to reproduce that algorithm.
            \item If the contribution is primarily a new model architecture, the paper should describe the architecture clearly and fully.
            \item If the contribution is a new model (e.g., a large language model), then there should either be a way to access this model for reproducing the results or a way to reproduce the model (e.g., with an open-source dataset or instructions for how to construct the dataset).
            \item We recognize that reproducibility may be tricky in some cases, in which case authors are welcome to describe the particular way they provide for reproducibility. In the case of closed-source models, it may be that access to the model is limited in some way (e.g., to registered users), but it should be possible for other researchers to have some path to reproducing or verifying the results.
        \end{enumerate}
    \end{itemize}

\item {\bf Open access to data and code}
    \item[] Question: Does the paper provide open access to the data and code, with sufficient instructions to faithfully reproduce the main experimental results, as described in supplemental material?
    \item[] Answer: \answerYes{} 
    \item[] Justification: We submit the code together with this paper. 
    \item[] Guidelines:
    \begin{itemize}
        \item The answer \answerNA{} means that paper does not include experiments requiring code.
        \item Please see the NeurIPS code and data submission guidelines (\url{https://neurips.cc/public/guides/CodeSubmissionPolicy}) for more details.
        \item While we encourage the release of code and data, we understand that this might not be possible, so \answerNo{} is an acceptable answer. Papers cannot be rejected simply for not including code, unless this is central to the contribution (e.g., for a new open-source benchmark).
        \item The instructions should contain the exact command and environment needed to run to reproduce the results. See the NeurIPS code and data submission guidelines (\url{https://neurips.cc/public/guides/CodeSubmissionPolicy}) for more details.
        \item The authors should provide instructions on data access and preparation, including how to access the raw data, preprocessed data, intermediate data, and generated data, etc.
        \item The authors should provide scripts to reproduce all experimental results for the new proposed method and baselines. If only a subset of experiments are reproducible, they should state which ones are omitted from the script and why.
        \item At submission time, to preserve anonymity, the authors should release anonymized versions (if applicable).
        \item Providing as much information as possible in supplemental material (appended to the paper) is recommended, but including URLs to data and code is permitted.
    \end{itemize}

\item {\bf Experimental setting/details}
    \item[] Question: Does the paper specify all the training and test details (e.g., data splits, hyperparameters, how they were chosen, type of optimizer) necessary to understand the results?
    \item[] Answer: \answerYes{} 
    \item[] Justification: We provide our experimental setting as well as the training and evaluating details in Section~\ref{sec:experiment_setup} and in Appendix~\ref{appendix:training_detail}, . 
    \item[] Guidelines:
    \begin{itemize}
        \item The answer \answerNA{} means that the paper does not include experiments.
        \item The experimental setting should be presented in the core of the paper to a level of detail that is necessary to appreciate the results and make sense of them.
        \item The full details can be provided either with the code, in appendix, or as supplemental material.
    \end{itemize}

\item {\bf Experiment statistical significance}
    \item[] Question: Does the paper report error bars suitably and correctly defined or other appropriate information about the statistical significance of the experiments?
    \item[] Answer: \answerNo{} 
    \item[] Justification: Due to the high computational cost, results are typically reported without error bars or statistical significance tests. To mitigate randomness, we perform multiple sampling runs in most of our evaluations to ensure the reliability of the results. 
    \item[] Guidelines:
    \begin{itemize}
        \item The answer \answerNA{} means that the paper does not include experiments.
        \item The authors should answer \answerYes{} if the results are accompanied by error bars, confidence intervals, or statistical significance tests, at least for the experiments that support the main claims of the paper.
        \item The factors of variability that the error bars are capturing should be clearly stated (for example, train/test split, initialization, random drawing of some parameter, or overall run with given experimental conditions).
        \item The method for calculating the error bars should be explained (closed form formula, call to a library function, bootstrap, etc.)
        \item The assumptions made should be given (e.g., Normally distributed errors).
        \item It should be clear whether the error bar is the standard deviation or the standard error of the mean.
        \item It is OK to report 1-sigma error bars, but one should state it. The authors should preferably report a 2-sigma error bar than state that they have a 96\% CI, if the hypothesis of Normality of errors is not verified.
        \item For asymmetric distributions, the authors should be careful not to show in tables or figures symmetric error bars that would yield results that are out of range (e.g., negative error rates).
        \item If error bars are reported in tables or plots, the authors should explain in the text how they were calculated and reference the corresponding figures or tables in the text.
    \end{itemize}

\item {\bf Experiments compute resources}
    \item[] Question: For each experiment, does the paper provide sufficient information on the computer resources (type of compute workers, memory, time of execution) needed to reproduce the experiments?
    \item[] Answer: \answerYes{} 
    \item[] Justification: We introduced the computational resource in Appendix~\ref{appendix:implementation}. 
    \item[] Guidelines:
    \begin{itemize}
        \item The answer \answerNA{} means that the paper does not include experiments.
        \item The paper should indicate the type of compute workers CPU or GPU, internal cluster, or cloud provider, including relevant memory and storage.
        \item The paper should provide the amount of compute required for each of the individual experimental runs as well as estimate the total compute. 
        \item The paper should disclose whether the full research project required more compute than the experiments reported in the paper (e.g., preliminary or failed experiments that didn't make it into the paper). 
    \end{itemize}
    
\item {\bf Code of ethics}
    \item[] Question: Does the research conducted in the paper conform, in every respect, with the NeurIPS Code of Ethics \url{https://neurips.cc/public/EthicsGuidelines}?
    \item[] Answer: \answerYes{} 
    \item[] Justification:  We confirm that our research complies with the NeurIPS Code of Ethics. 
    \item[] Guidelines: 
    \begin{itemize}
        \item The answer \answerNA{} means that the authors have not reviewed the NeurIPS Code of Ethics.
        \item If the authors answer \answerNo, they should explain the special circumstances that require a deviation from the Code of Ethics.
        \item The authors should make sure to preserve anonymity (e.g., if there is a special consideration due to laws or regulations in their jurisdiction).
    \end{itemize}

\item {\bf Broader impacts}
    \item[] Question: Does the paper discuss both potential positive societal impacts and negative societal impacts of the work performed?
    \item[] Answer: \answerYes{} 
    \item[] Justification:  We discussed our potential impacts in~\ref{appendix:social_risks}

    \item[] Guidelines:
    \begin{itemize}
        \item The answer \answerNA{} means that there is no societal impact of the work performed.
        \item If the authors answer \answerNA{} or \answerNo, they should explain why their work has no societal impact or why the paper does not address societal impact.
        \item Examples of negative societal impacts include potential malicious or unintended uses (e.g., disinformation, generating fake profiles, surveillance), fairness considerations (e.g., deployment of technologies that could make decisions that unfairly impact specific groups), privacy considerations, and security considerations.
        \item The conference expects that many papers will be foundational research and not tied to particular applications, let alone deployments. However, if there is a direct path to any negative applications, the authors should point it out. For example, it is legitimate to point out that an improvement in the quality of generative models could be used to generate Deepfakes for disinformation. On the other hand, it is not needed to point out that a generic algorithm for optimizing neural networks could enable people to train models that generate Deepfakes faster.
        \item The authors should consider possible harms that could arise when the technology is being used as intended and functioning correctly, harms that could arise when the technology is being used as intended but gives incorrect results, and harms following from (intentional or unintentional) misuse of the technology.
        \item If there are negative societal impacts, the authors could also discuss possible mitigation strategies (e.g., gated release of models, providing defenses in addition to attacks, mechanisms for monitoring misuse, mechanisms to monitor how a system learns from feedback over time, improving the efficiency and accessibility of ML).
    \end{itemize}
    
\item {\bf Safeguards}
    \item[] Question: Does the paper describe safeguards that have been put in place for responsible release of data or models that have a high risk for misuse (e.g., pre-trained language models, image generators, or scraped datasets)?
    \item[] Answer: \answerNA{} 
    \item[] Justification: Our work does not involve methods that pose significant risks. 
    \item[] Guidelines:
    \begin{itemize}
        \item The answer \answerNA{} means that the paper poses no such risks.
        \item Released models that have a high risk for misuse or dual-use should be released with necessary safeguards to allow for controlled use of the model, for example by requiring that users adhere to usage guidelines or restrictions to access the model or implementing safety filters. 
        \item Datasets that have been scraped from the Internet could pose safety risks. The authors should describe how they avoided releasing unsafe images.
        \item We recognize that providing effective safeguards is challenging, and many papers do not require this, but we encourage authors to take this into account and make a best faith effort.
    \end{itemize}

\item {\bf Licenses for existing assets}
    \item[] Question: Are the creators or original owners of assets (e.g., code, data, models), used in the paper, properly credited and are the license and terms of use explicitly mentioned and properly respected?
    \item[] Answer: \answerYes{} 
    \item[] Justification: We properly cite all datasets, models, and codebases used in this work.
The corresponding licenses and terms of use are respected.
Publicly available assets are referenced in the main paper and supplementary material where appropriate. 
    \item[] Guidelines:
    \begin{itemize}
        \item The answer \answerNA{} means that the paper does not use existing assets.
        \item The authors should cite the original paper that produced the code package or dataset.
        \item The authors should state which version of the asset is used and, if possible, include a URL.
        \item The name of the license (e.g., CC-BY 4.0) should be included for each asset.
        \item For scraped data from a particular source (e.g., website), the copyright and terms of service of that source should be provided.
        \item If assets are released, the license, copyright information, and terms of use in the package should be provided. For popular datasets, \url{paperswithcode.com/datasets} has curated licenses for some datasets. Their licensing guide can help determine the license of a dataset.
        \item For existing datasets that are re-packaged, both the original license and the license of the derived asset (if it has changed) should be provided.
        \item If this information is not available online, the authors are encouraged to reach out to the asset's creators.
    \end{itemize}

\item {\bf New assets}
    \item[] Question: Are new assets introduced in the paper well documented and is the documentation provided alongside the assets?
    \item[] Answer: \answerNA{} 
    \item[] Justification: We do not release new assets. 
    \item[] Guidelines:
    \begin{itemize}
        \item The answer \answerNA{} means that the paper does not release new assets.
        \item Researchers should communicate the details of the dataset\slash code\slash model as part of their submissions via structured templates. This includes details about training, license, limitations, etc. 
        \item The paper should discuss whether and how consent was obtained from people whose asset is used.
        \item At submission time, remember to anonymize your assets (if applicable). You can either create an anonymized URL or include an anonymized zip file.
    \end{itemize}

\item {\bf Crowdsourcing and research with human subjects}
    \item[] Question: For crowdsourcing experiments and research with human subjects, does the paper include the full text of instructions given to participants and screenshots, if applicable, as well as details about compensation (if any)? 
    \item[] Answer: \answerNA{} 
    \item[] Justification: We do not involve crowdsourcing nor research with human subjects. 
    \item[] Guidelines:
    \begin{itemize}
        \item The answer \answerNA{} means that the paper does not involve crowdsourcing nor research with human subjects.
        \item Including this information in the supplemental material is fine, but if the main contribution of the paper involves human subjects, then as much detail as possible should be included in the main paper. 
        \item According to the NeurIPS Code of Ethics, workers involved in data collection, curation, or other labor should be paid at least the minimum wage in the country of the data collector. 
    \end{itemize}

\item {\bf Institutional review board (IRB) approvals or equivalent for research with human subjects}
    \item[] Question: Does the paper describe potential risks incurred by study participants, whether such risks were disclosed to the subjects, and whether Institutional Review Board (IRB) approvals (or an equivalent approval/review based on the requirements of your country or institution) were obtained?
    \item[] Answer: \answerNA{} 
    \item[] Justification: We do not involve crowdsourcing nor research with human subjects.
    \item[] Guidelines:
    \begin{itemize}
        \item The answer \answerNA{} means that the paper does not involve crowdsourcing nor research with human subjects.
        \item Depending on the country in which research is conducted, IRB approval (or equivalent) may be required for any human subjects research. If you obtained IRB approval, you should clearly state this in the paper. 
        \item We recognize that the procedures for this may vary significantly between institutions and locations, and we expect authors to adhere to the NeurIPS Code of Ethics and the guidelines for their institution. 
        \item For initial submissions, do not include any information that would break anonymity (if applicable), such as the institution conducting the review.
    \end{itemize}

\item {\bf Declaration of LLM usage}
    \item[] Question: Does the paper describe the usage of LLMs if it is an important, original, or non-standard component of the core methods in this research? Note that if the LLM is used only for writing, editing, or formatting purposes and does \emph{not} impact the core methodology, scientific rigor, or originality of the research, declaration is not required.
    \item[] Answer:  \answerNA{} 
    \item[] Justification: the core method development in this research does not involve LLMs as any important, original, or non-standard components.
    \item[] Guidelines: 
    \begin{itemize}
        \item The answer \answerNA{} means that the core method development in this research does not involve LLMs as any important, original, or non-standard components.
        \item Please refer to our LLM policy in the NeurIPS handbook for what should or should not be described.
    \end{itemize}

\end{enumerate}

\end{document}